\documentclass[sigconf, nonacm]{acmart}
\usepackage{dsbda}

\AtBeginDocument{%
  \providecommand\BibTeX{{%
    \normalfont B\kern-0.5em{\scshape i\kern-0.25em b}\kern-0.8em\TeX}}}

\setcopyright{acmcopyright}
\copyrightyear{2021}
\acmYear{2021}

\usepackage{makecell}

\usepackage{multirow}

\usepackage{subcaption}
\usepackage{xcolor} 
\usepackage{todonotes}
\usepackage[nolist]{acronym}
\newcommand{\HeterSUMGraph}{H\textsc{eter}SUMG\textsc{raph} }
\usepackage{bm}

\begin{acronym}
    \acro{MDS}{Multi-document Summarization}
    \acro{AWD}{Attention Weight Distribution}
    \acro{GAT}{Graph Attention Network}
    \acro{ADG}{Approximate Discourse Graph}
    \acro{GCN}{Graph Convolution Network}
    \acro{KG}{Knowledge Graph}
\end{acronym}

\begin{document}

\author{M. Lautaro Hickmann}
\affiliation{
\institution{Ulm University}}
\email{lautaro.hickmann@uni-ulm.de}

\author{Fabian Wurzberger}
\affiliation{
\institution{Ulm University}}
\email{fabian.wurzberger@uni-ulm.de}

\author{Megi Hoxhalli}
\affiliation{
\institution{Ulm University}}
\email{megi.hoxhalli@uni-ulm.de}

\author{Arne Lochner}
\affiliation{
\institution{Ulm University}}
\email{arne.lochner@uni-ulm.de}

\author{Jessica Töllich}
\authornote{Author is contributing Conceptualization, Writing - Review \& Editing, and Supervision.
Statement is based on the Contributor Roles Taxonomy, see: 
\url{http://credit.niso.org/}
}
\affiliation{
\institution{Ulm University}}
\email{jessica.toellich@uni-ulm.de}

\author{Ansgar Scherp}
\authornote{Author is contributing Conceptualization, Writing - Review \& Editing, and Supervision.
Statement is based on the Contributor Roles Taxonomy, see: 
\url{http://credit.niso.org/}
}
\affiliation{
\institution{Ulm University}}
\email{ansgar.scherp@uni-ulm.de}


\title{Analysis of GraphSum's Attention Weights to Improve the Explainability of Multi-Document Summarization}

\begin{abstract}
Modern multi-document summarization (MDS) methods are based on transformer architectures. 
They generate state of the art summaries, but lack explainability.
We focus on graph-based transformer models for MDS as they gained recent popularity.
We aim to improve the explainability of the graph-based MDS by analyzing their attention weights.
In a graph-based MDS such as GraphSum, vertices represent the textual units, while the edges form some similarity graph over the units. 
We compare GraphSum's performance utilizing different textual units, i.\,e., sentences versus paragraphs, on two news benchmark datasets, namely WikiSum and MultiNews.
Our experiments show that paragraph-level representations provide the best summarization performance.
Thus, we subsequently focus on analyzing the paragraph-level attention weights of GraphSum's multi-heads and decoding layers in order to improve the explainability of a transformer-based MDS model.
As a reference metric, we calculate the ROUGE scores between the input paragraphs and each sentence in the generated summary, which indicate source origin information via text similarity.
We observe a high correlation between the attention weights and this reference metric, especially on the the later decoding layers of the transformer architecture.
Finally, we investigate if the generated summaries follow a pattern of positional bias by extracting which paragraph provided the most information for each generated summary.
Our results show that there is a high correlation between the position in the summary and the source origin. 
\end{abstract}

\maketitle

\section{Introduction}
\label{sec:intro}

\ac{MDS} refers to the task of providing a concise representation of multiple documents with overlapping textual content~\cite{lin2002single}.
Besides transformer-based models for summarization~\cite{liu2019hierarchical, liu2019text}, approaches on graph neural networks and specifically using knowledge graphs have gained popularity~\cite{li2020leveraging,huang2020knowledge, yasunaga-etal-2017-graph, xu2020discourseaware} while more recent models combine graph neural networks with a transformer architecture \cite{li2020leveraging, huang2020knowledge}.

Generally, we distinguish extractive and abstractive MDS.
In \textit{extractive \ac{MDS}}, a model is trained to select the most relevant input sentences or paragraphs in order to create an optimal summary with regard to salience and coherence~\cite{yasunaga-etal-2017-graph}. 
In comparison, \textit{abstractive \ac{MDS}} is able to generate new texts which are not present in the input documents, which improve the summary quality~\cite{huang2020knowledge}. 
For extractive \ac{MDS}, a graph-based representation can help to detect salient vertices, so that they are  included in the generated summary to cover all important information. 
In abstractive \ac{MDS}, salient vertices can guide the summarization process via attention mechanisms of the transformer architecture. 
To this end, the text from multiple documents is represented as independent, so-called \textit{textual units}. 
The graph's vertices represent the textual units such as paragraphs and sentences, while the edges model a semantic relationship between the units like a text-based cosine similarity.
Different textual units have been used for graph-based text summarization in the past, including discourse relations~\cite{huang2020knowledge, xu2020discourseaware}, sentences ~\cite{christensen-etal-2013-towards, yasunaga-etal-2017-graph}, and paragraphs ~\cite{li2020leveraging}.

Recently, \citet{li2020leveraging} propsed GraphSum, a state of the art graph-based \ac{MDS}, which utilizes a graph structure based on paragraphs as textual units to improve the transformer architecture and guide the summary generation process~\cite{li2020leveraging}. 
Paragraphs are assumed to divide text into contiguous topics. 
Leveraging inter-paragraph relations can provide the model additional information for detecting contextual relations between topics. 
The GraphSum model showed substantial improvements compared to strong \ac{MDS} baselines such as LEAD-3 and LexRank ~\cite{erkan2004lexrank}. 

The authors of GraphSum~\cite{li2020leveraging} did considered paragraph-level textual units as vertices for their graph structure. 
Also \citeauthor{huang2020knowledge}\cite{huang2020knowledge} and others used paragraph-level representations for the graph-based MDS task.
Other works for \ac{MDS} considered sentence-level textual units for the summarization tasks such as~\cite{christensen-etal-2013-towards,yasunaga-etal-2017-graph, wang2020heterogeneous}.
However, to the best of our knowledge, there is no graph-based \ac{MDS} model that has used representation of textual units on both levels, \ie paragraph-level and sentence-level, on the same dataset.
Thus, in a first step, we investigate if a finer-grained text representation with textual units on sentence level improves the results.
We focus on using GraphSum as state of the art model to contrast the summarization results of a representation on paragraph level vs sentence level.
Our rationale is that with sentences as textual units, the graph structure represents inter-sentence relations, which may provide more detailed information within topics and thus may improve the results. 

Based on the first step, we are interested whether a graph-based \ac{MDS} model extracts information from textual units based on their position in the content.
It would mean that the summarization model is capable of picking up this bias from the input data.
Such a \textit{source origin analysis} is particularly interesting for datasets where the gold standard is biased towards selecting information from the first sentences.
In example, for \ac{MDS} tasks within the news domain, Kedzie et al.~\cite{kedzie2019content} and Jung et al. ~\cite{jung2019earlier} suggest that news summarizations follows a \emph{positional bias} pattern.
In the news domain, the information from the first sentences of the input documents is the most relevant. 
This is the main reason why the simple, \textit{extractive} summarization model LEAD-3, where the first 3 sentences are selected to generate a summary, performs well as an model. 
However, the LEAD-3 model is biased towards the news domain.
It exploits the specific knowledge about how journalists write their articles.
In abstractive \ac{MDS}, such as GraphSum, new sentences are generated and therefore no explainable source origin information can be directly extracted from the generated summaries.
However, abstractive transformer-based \ac{MDS} models often use positional encodings ~\cite{vaswani2017attention} to make use of the sequential structure of the data. 
Thus, in the second step we analyze the attention weights in the GraphSum model to improve the understanding of abstractive \ac{MDS}.

In summary, we investigate the following two research questions:
\begin{enumerate}
\item We construct similarity graphs on sentence-level and para\-graph-level in order to guide a graph-transformer-based \ac{MDS} model~\cite{liu2018generating}.
We investigate which of the aforementioned textual units are most beneficial as vertices in a similarity graph to capture context and concepts of the input documents and therefore guide the \ac{MDS} process most effectively. 

\item We analyze attention mechanisms in an abstractive graph-transformer-based \ac{MDS} model~\cite{liu2018generating} in order to detect which specific textual units provide the most valuable information for a generated summary.
We use the two benchmark datasets MultiNews and WikiSum.
This shall shed light onto the decisions made by the \ac{MDS} regarding the positions of the text units within the larger documents, \ie make more explainable from which part of the input the resulting summaries are generated.
 \end{enumerate}

Thus, overall the contribution of this work is to better understand how similarity graphs guide the summarization process. 
%
Regarding the first step, our experiments show that sentence-level representations of textual units in GraphSum are outperformed by paragraph-level representations for \ac{MDS} on the MultiNews dataset. 
We focus on the news domain, and particularly the benchmark dataset MultiNews, in preparation to the second step on the analysis of the attention weights as motivated above.
Regarding the results of the second step, we analyze the attention weights of GraphSum on two benchmark datasets, namely MultiNews and WikiSum.
Our results indicate that the attention weights of later decoding layers of the transformer architecture can provide source origin information. 
This suggests that attention weights can be used in further work to analyze if positional information (\eg positional encoding of textual units) in abstractive transformer-based \ac{MDS} leads to the side effect, that trained models learn a positional bias as mentioned previously. 

The remainder of the article is organized as follows.
Below we discuss related work on \ac{MDS}, with specific focus on methods that leverage graph structures.
In Section~\ref{sec:research-methods}, we introduce the methods and data pre-processing two answer the two research questions.
In Section~\ref{sec:experimental}, we describe the experimental apparatus.
Section \ref{sec:results} presents the experimental results.
We discuss the results in Section \ref{sec:discussion}, before we conclude.

\section{Related Work}
\label{sec:related-work}
\label{sec:related-work:gbmd}

\acf{MDS} can be viewed as a sequence-to-sequence task~ \cite{li2020leveraging}, where a set of multiple input documents is transformed into a single text summary. 
Most abstractive \ac{MDS} models were based on an auto-regressive encoder-decoder architecture, where the encoder maps an input sequence to a hidden representation and the decoder transforms the hidden representation into another sequence in a token-by-token manner \cite{vaswani2017attention}. 
More elaborate approaches further included an attention mechanism in the encoder-decoder architecture, which improves the models when using long sequences \cite{bahdanau15attention,vaswani2017attention}.

Recently graph structures were used to improve the performance of extractive and abstractive \ac{MDS} models by guiding the summarization process implicitly and explicitly. Similarity and discourse graphs have been used for \ac{MDS} ~\cite{li2020leveraging}. 
Especially similarity graphs, which are able to capture lexical or topic relations between textual units, improve the quality of the generated summaries ~\cite{li2020leveraging}.
 Below, we review the state of the art in graph-based extractive and graph-based abstractive  \ac{MDS} approaches.

\subsection{Extractive Graph-based Summarization}
\label{sec:related-work:gbmd:extractive}
An early work from 2013 by Christensen et al. ~\cite{christensen-etal-2013-towards} proposed to leverage graph-structures for extractive text summarization in order to improve the coherence of the generated summary. 
They constructed a directed discourse graph with input sentences as vertices, where the edges between two vertices represent constraints for sentence ordering to achieve a coherent summary. 
The process of retrieving edge weights for this discourse graph was based on heuristic metrics such as deverbal noun references, event-entity continuation, discourse markers, inferred edges, and co-referent mentions.
These metrics were used to construct the \ac{ADG}. 
All possible candidate summaries for the input documents were evaluated based on a joint objective function, where coherence and salience were a soft constraint.
The \ac{ADG} enabled the \ac{MDS} model to estimate the coherence score of a candidate summary by analyzing the edge weights of successive summary sentences. 
A linear regression model was trained based on ROGUE scores to predict the salience score of a candidate summary.
The candidate summary, which maximized this function was returned as the generated summary.

As follow up work, \citeauthor{yasunaga-etal-2017-graph} ~\cite{yasunaga-etal-2017-graph} improved the creation process of the \ac{ADG} in 2017. 
While Christensen et al. ~\cite{christensen-etal-2013-towards} used the graph structure only implicitly for constraint purposes, \citeauthor{yasunaga-etal-2017-graph} ~\cite{yasunaga-etal-2017-graph}
used the graph structure explicitly for a \ac{GCN} ~\cite{kipf2016semi}.
\citeauthor{yasunaga-etal-2017-graph} noticed that the absolute value of the edge weights in \citeauthor{christensen-etal-2013-towards}'s \ac{ADG} lacked diversity, because they were only incremented by discrete values and many edge weights were equal to one. 
Thus, the edge weights did not provide additional information for the summarization beyond the simple knowledge about the presence of a relation between two sentences. 
To overcome this issue, \citeauthor{yasunaga-etal-2017-graph} first transformed the edge weights for each sentence via sentence personalization and afterwards normalized over all incoming edges. 
This resulted in the Personalized Discourse Graph (PDG), which reflected macro-level sentence features.
They were helpful to estimate salience information. 
In the experiments, PDG performed better than \ac{ADG} with regards to ROUGE scores, learning curves, and computation costs. 
Additionally, the node degree in PDG has had a higher correlation to the salience score estimated by the linear regression than its counterpart \ac{ADG}.

More recently, \citeauthor{wang2020heterogeneous} \cite{wang2020heterogeneous}  presented a new architecture named ~\HeterSUMGraph for extractive summarization based upon a heterogenous graph-based neural network, built by nodes of varying granularity. 
These nodes were then used to enrich the cross sentence relations, making it flexible to use in single-document and multi-document summarization.
The model consists of three parts: graph initializer, heterogenous graph layer, and sentence selector. 
To create the sentence nodes, they used a concatenation of a CNN and BiLSTM, and for the relations between them, they made use of TF-IDF. 
For the heterogenous graph layer, \HeterSUMGraph used \ac{GAT}, and simple feed-forward neural network in an iterative process. 
Finally, in the sentence selecting process, trigram blocking was used to score and discard redundant sentences. 

\subsection{Abstractive Graph-based Summarization}
\label{sec:related-work:gbmd:abstractive}
While Christensen et al.~\cite{christensen-etal-2013-towards} and Yasanaga et al.~\cite{yasunaga-etal-2017-graph} already leveraged discourse graphs in an implicit and explicit manner, Li et al. ~\cite{li2020leveraging} extended this idea by using similarity graphs based on TF-IDF as well as topic graphs in a transformer-based architecture~\cite{vaswani2017attention}. 
The graph structure of the input data was used to guide the attention mechanism within the encoder and decoder of the transformer-like architecture.
This allowed the adjacency-matrix to explicitly influence the encoder and decoder part of the summarizer. 
The resulting system was called \emph{GraphSum}.
The graph encoding layers applied graph-informed self-attention onto the input data, where the attention weights are shifted by the graph-information.
The decoder within the GraphSum model contained a Global Graph Attention mechanism. 
This mechanism introduced a global context vector, which is a weighted sum of the central textual unit vectors. 
In addition, the decoder contained a local normalized attention, in which each input token is attended separately. The inference process of GraphSum was extended by a beam search in order to improve the quality of generated summaries.
For evaluation of GraphSum, \citeauthor{li2020leveraging}~ \cite{li2020leveraging} used the WikiSum-Dataset as well as the MultiNews-Dataset. 
For both datasets, paragraphs acted as the central textual units for the graph-structure.
Sentence-level representations have not been considered.

Another approach is ASGARD by \citeauthor{huang2020knowledge} \cite{huang2020knowledge}.
ASGARD used \acp{GAT} to capture global context based on \acp{KG}, which are constructed by an openIE model. 
The authors argued that topics are naturally split by paragraphs, therefore they also employed their \ac{GAT} encoding on paragraph level by sampling subgraphs from the \ac{KG}. 
Furthermore, they also followed the commonly used encoder-decoder architecture. Additionally, they used graph attention and token level attention in the decoder part. 
Besides next token prediction, \citeauthor{huang2020knowledge} introduced the additional task of node salience labeling, where the model is trained to predict a salience score for each node. 
Those scores were based on the occurrences of nodes in the gold standard and were used to filter nodes in the graph encoder. 
In addition, to optimize the cross entropy loss for the next token prediction, \citeauthor{huang2020knowledge} further assumed that informative summaries yield better accuracy on question answering tasks. 
Accordingly, they defined a reinforcement objective based on training a question answering model on the generated summaries.

\subsection{Summary}
\label{sec:related-work:summary}
In contrast to existing works, we investigate the influence of different textual units \ie using sentence-level vs. paragraph-level textual units, for training the same abstractive graph-based \ac{MDS} model. 
In addition, we investigate the possibility to utilize attention weights of a transformer-based decoder in order to extract source origin information of a generated summary. 

Our analysis is based on the  GraphSum~\cite{li2020leveraging} model, which delivers state-of-the-art results for graph-transformer-based \ac{MDS} architectures, enabling a source origin analysis via attention weights.
GraphSum is extended by pre-processing the input data to represent sentence-level and paragraph-level textual units.
In addition, we extract the generated global attention weights of a trained GraphSum model in order to analyze the possibility of leveraging attention weights to improve the explainability of abstractive \ac{MDS}.

\section{Extension of the GraphSum MDS}
\label{sec:research-methods}

We base on the state of the art abstractive graph-based transformer GraphSum from Li et al.~\cite{li2020leveraging}.\footnote{We use the code provided by the authors, which is available here: \url{https://github.com/PaddlePaddle/Research/tree/master/NLP/ACL2020-GraphSum}}
GraphSum uses a graph-based representation of the input paragraphs for the encoder and decoder part of the transformer architecture.
Thus, GraphSum is optimally suited to be extended to incorporate a sentence-level representation, to answer the first research questions. 
Additionally, the extraction of attention weights enables the possibility to analyze source origin relations, to answer the second research question.

Below, we first describe the modification of GraphSum to support both paragraph-level and sentence-level text unit representation.
Subsequently, we explain how we extract the attention weights for obtaining the positional information to make the summaries more transparent.

\subsection{Modifying GraphSum for Supporting Sentence-level Text Representations}
\label{sec:research-methods:GraphSum}

The input data of the GraphSum model is tokenized using a language model, namely \emph{sentencepiece}~\cite{kudo2018sentencepiece}. Then truncating and padding are applied to each input textual unit, as done by GraphSum ~\citet{li2020leveraging}. This results in each input set of multi-documents containing a fixed number of textual units and tokens within each unit. The positional information of each unit is preserved by utilizing positional encoding. This basic pre-processing transforms the input data into a sequence of tokens. 
This allows us to train GraphSum with minor extensions on both sentence-level and paragraph-level document representations.

In order to compare the results of the paragraph-level model and sentence-level model, we need to make sure that the sentence-level models receives the same number of tokens compared to he paragraph-level model. 
While the total input to the model on sentence-level has the same length, the actual sequence of tokens can slightly differ between the two levels of textual units due to difference in the truncation and padding procedure.
%
The already pre-processed paragraph-level dataset \cite{li2020leveraging} is split into sentence with the help of a tokenizer, namely \emph{sentencepiece}~\cite{kudo2018sentencepiece}. This will result in an increase of textual units, because each paragraph must contain at least one sentence. The newly obtained textual units differ with regard to the number of tokens. Therefore, the data is truncated and padded to a fixed number of input sentences and a fixed number of tokens for each sentence. 
The size of the data immensely increases when applying padding on sentence-level text representation (as it is done for the paragraph-level representation), due to the higher number of textual units.
The padding is an important factor, as it defines the maximal number of tokens a textual unit can have. 
Therefore, for sentence-level representation the maximal number of tokens per sentence, \ie per textual unit is reduced, in order to increase the maximal total number of textual units. 
Afterwards we build a TF-IDF graph-structure to capture similarities between the newly obtained sentences. The already pre-processed paragraph-level dataset already contains a TF-IDF graph-structure. 
Therefore, no further pre-processing is necessary for the paragraph-level model. 
%

\subsection{Extracting the Source Origin via GraphSum's Attention Weights}
\label{sec:research-methods:orign}

GraphSum has in total 8 decoding layers with 8 multi-heads each.
A detailed description of the architecture can be found in~\cite{li2020leveraging}. 
Below we focus on explaining the global graph attention as we use this information for the analysis of the source origin information, \ie understanding the position of the paragraphs that most influence the summaries  generated by GraphSum.
 
\begin{figure}[h]
    \includegraphics[width=\columnwidth]{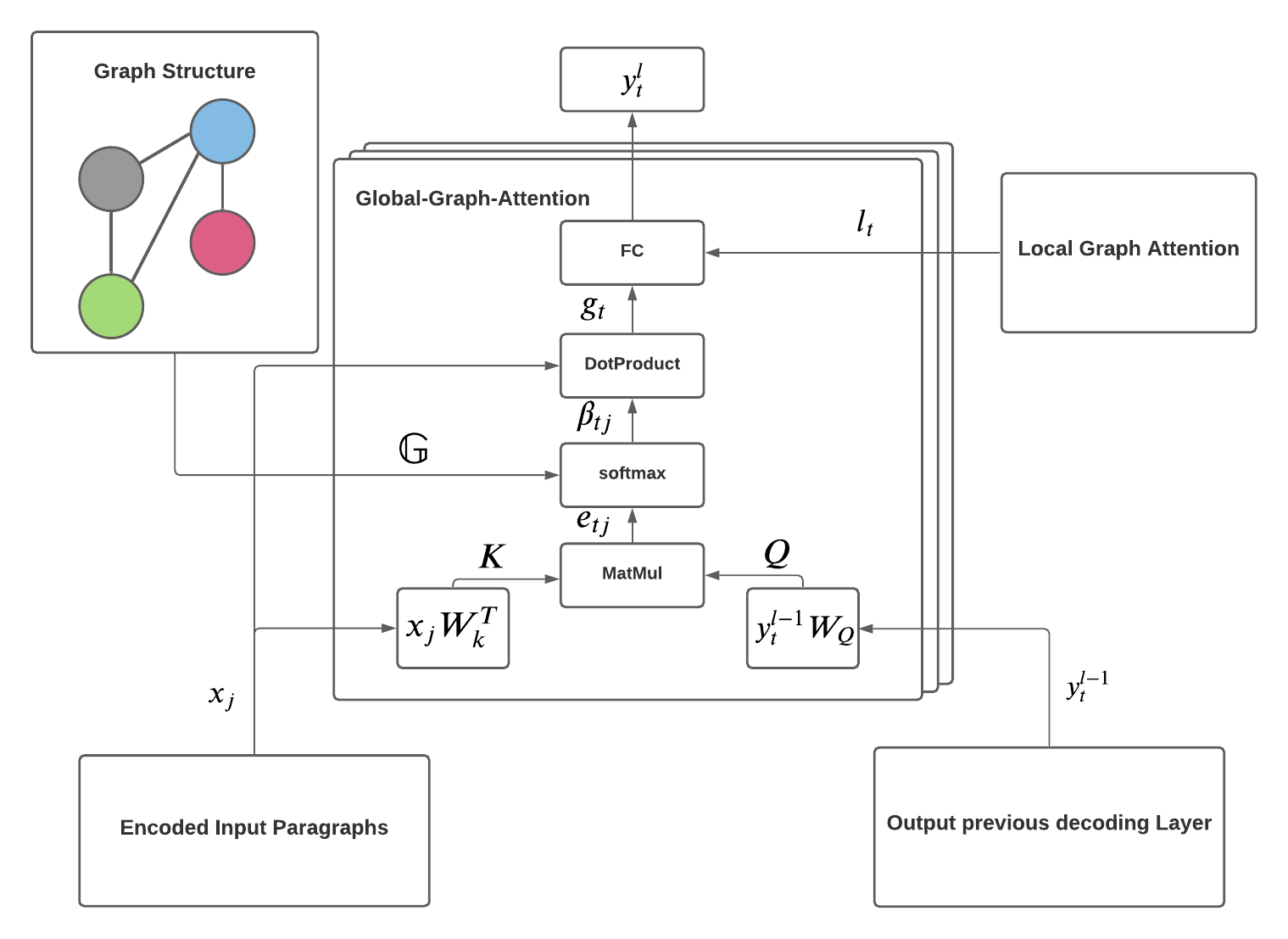}
    \centering
    \caption{Global graph attention mechanism within a single decoding layer. 
    Multi-Heads are indicated by the stacked boxes (center).}
    \label{fig:global-graph-attention}
\end{figure}

The global graph attention as described by \citeauthor{li2020leveraging} \cite{li2020leveraging} is illustrated in \autoref{fig:global-graph-attention}.
The global context vector $\vg_t$ within a decoding layer $l$ and the summary token at position $t$ is obtained by calculating a weighted sum of the attention distribution $\beta_{tj}$ and the encoded paragraphs' vectors $\vx_j, j\in \{1, \dots, L\}$, where $L$ denotes the number of input paragraphs:

\begin{align}
    \vg_t = \sum_{j=1}^L \beta_{tj} \vx_j
    \label{eq:attention}
\end{align}

The \ac{AWD} of decoding layer $\vl$ is calculated by  predicting the central paragraph $\vs_t$ for the attended token $\vy_t^{l-1}$ of the previous decoding layer. 
At each step of the auto-regressive decoder architecture, the central paragraph $\vs_t$ represents the paragraph, which is at central position for the next predicted token.
The central paragraph $\vs_t$ helps to guide the model to attend paragraphs, which are similar to $\vx_{s_t}$ according to the graph structure $\mathbb{G}$. 
This central position is obtained by transforming the attended token $\vy_t^{l-1}$ by a two-layered feed forward network. 
Note that the central paragraph is solely based on $\vy_t^{l-1}$, which corresponds to the query token of the current attention mechanism. 

Afterwards, the attention weights between encoded paragraph $\vx_j$ and token vector $\vy_t$ are shifted by leveraging graph information between the central paragraph and paragraph $\vx_j$. 
These shifted attention weights are then transformed into a probability distribution by applying a softmax function resulting in the \ac{AWD} denoted by $\beta$:

\begin{align}
        \beta_{tj} = \softmax(\ve_{tj} - \frac{(1 - \mathbb{G}[s_t][j]^2)}{2 \sigma^2})
\end{align}

The scaling factor $\sigma$ denotes the influence of the graph structure on the attention weights~\cite{li2020leveraging}.
The unscaled attention weights $\ve_{tj}$ between input paragraph $\vx_j$ and summary token $\vy_t^{l-1}$ are calculated according to \citet{vaswani2017attention}, where $d_{head}$ denotes the number of multi-heads:

\begin{align}
    \ve_{tj} = \frac{(\vy_t^{l-1} W_Q)(\vx_j W_k)^T}{\sqrt{d_{head}}}
\end{align}

After training the GraphSum model as described in \cite{li2020leveraging}, we generate summaries for each set of multi-documents from the test dataset and extract the \ac{AWD} for each generated token, multi-head, decoding layer, and beam.
This results in a tensor
$\tA \in \mathbb{R}^{bs \times sl \times dl \times mh \times L}$ for a single generated summary, where $bs$ is the number of beams, $sl$ the length of the generated summary, $dl$ is the number of decoding layers, $mh$ the number of multi-heads and $L$ denoting the number of input paragraphs.
Since GraphSum uses beam search for its inference process, $\tA$ needs to be transformed by a beam search decoder in order to obtain the correct \ac{AWD} at each step for the generated summaries.
    
As the auto-regressive decoder architecture generates a single token with corresponding \ac{AWD} at each step, this information can only be used to provide source origin information for each token individually.
But the source origin information of a single token within a generated summary might not be interesting. Instead it may be desirable to obtain source origin information about a sequence of tokens (\eg sentences).
For this purpose we split the generated summaries into the individual, generated sentences according to a special token which indicates the end of a sentence~\cite{kudo-richardson-2018-sentencepiece}.
This information is then used to aggregate the token-level information of $\tA$ to sentence-level by applying an aggregation method (\eg mean or median) resulting in $\tA'$.
Thus, the \ac{AWD} values of a single paragraph $\vx_j$ are an aggregation over all tokens of each generated sentence.

To analyze the possibility of leveraging attention weights for source origin information, we compute the correlation between $\tA'$ and a source origin metric $\tR$, which is based on the similarity score between the generated summary and the input paragraphs.
We use ROUGE scores, as described in Section~\ref{sec:experimental:measures}.

\section{Experimental Apparatus}
\label{sec:experimental}

\subsection{Datasets}
\label{sec:experimental:datasets}
We use two datasets, MultiNews and WikiSum, which are used as benchmarks in evaluating \ac{MDS} models.
MultiNews~\cite{fabbri2019multinews} contains human written summaries from professionals and includes links to the original articles. 
The articles were taken from \url{newser.com} and comprises about $60,000$ documents.
We use the pre-processed data from \citeauthor{li2020leveraging} \cite{li2020leveraging}.  

WikiSum is a \ac{MDS} dataset based on references found in Wikipedia articles and Google searches of the Wikipedia articles' titles~\cite{liu2018generating}. 
The gold standard summary is the corresponding Wikipedia article itself. 
WikiSum contains about $2.3$ million Wikipedia articles. 
We will use the ranked version of the WikiSum dataset, which was suggested by Liu et al. ~\cite{liu2019hierarchical}.  
The ranked dataset contains the top-$40$ paragraphs, which were predicted by a paragraph ranker to be most relevant as described by \citeauthor{liu2019hierarchical}~\cite{liu2019hierarchical}. 

We use the MultiNews dataset for sentences vs. paragraphs comparison.
Due to resource restriction, we do not consider the Wiki\-Sum dataset for the sentences vs. paragraphs comparison.
We use both datasets, \ie MultiNews and WikiSum, for analyzing the source origin information via attention weights.

\subsection{Procedure}
\label{sec:experimental:procedure}

\paragraph{Summarization using sentences vs. paragraphs.}
\label{sec:experimental:procedure:sentence-paragraph}
Regarding the comparison of sentence-level vs. paragraph-level representation of the input documents, we train our GraphSum models on batch sizes of $3,000$. 
The batch size represents the number of target tokens. 
Thus, we use the same batch size for both representation to make a fair comparison of the results.
Additionally, for the paragraph-level model, we also use a batch size of $4,000$, which is not possible anymore for the sentence-level model due to the higher memory requirements. 
We conduct this comparison on the MultiNews dataset.
We split the dataset into $44,927$ 
samples for training, $5,622$ for validation, and $5,622$ for testing.

\begin{figure}[h]
    \includegraphics[width=\columnwidth]{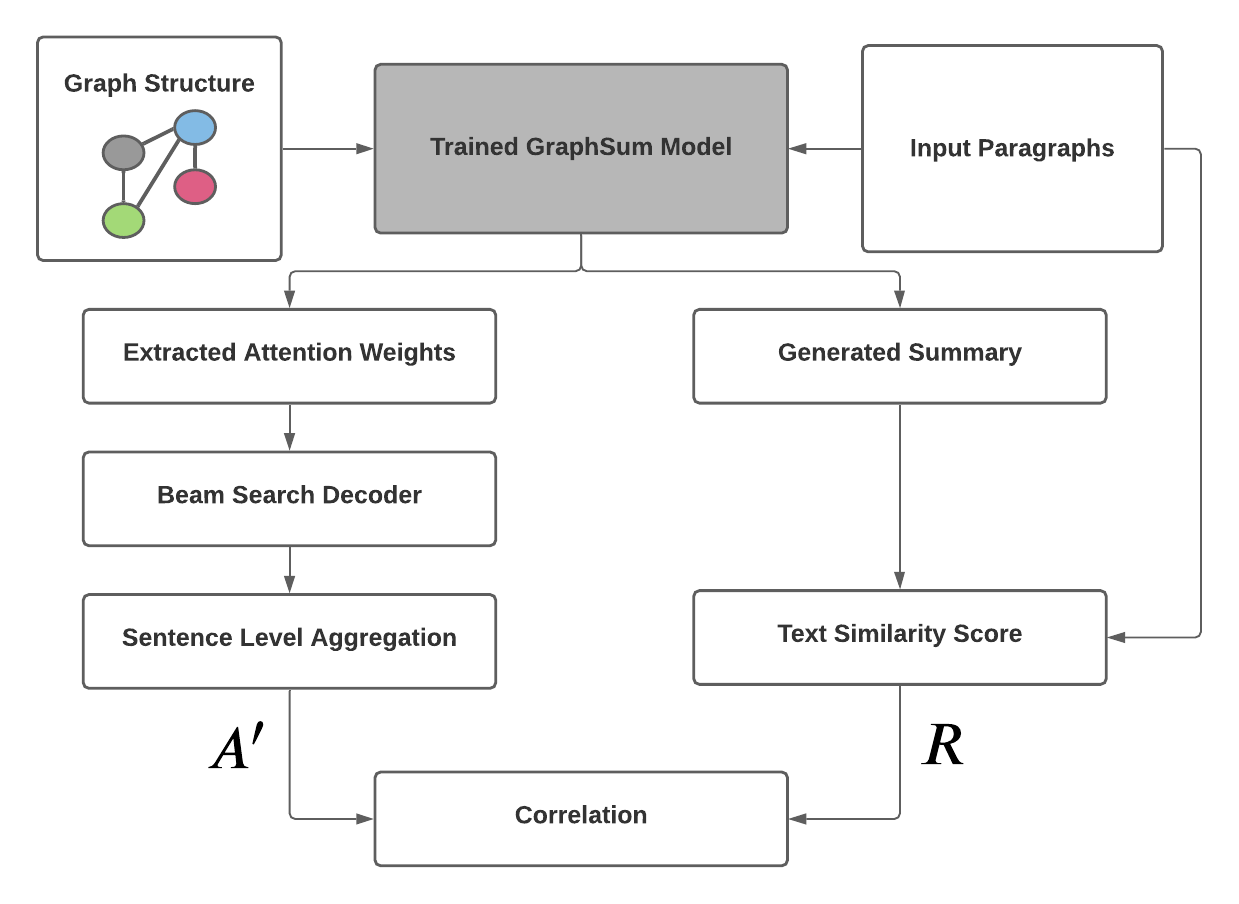}
    \centering
    \caption{Processing pipeline to calculate correlation between aggregated \ac{AWD} $\tA'$ and reference text similarity metric $\tR$.}
    \label{fig:rq2-pipeline}
\end{figure}

\paragraph{Source origin information via attentions weights.}
\label{sec:experimental:procedure:source-origin}
Regarding the second research question, \ie analyzing the possible correlation between source origin information and the attention weights, we use the pre-trained paragraph-level GraphSum model provided by \citet{li2020leveraging}. 
In the inference process on the test data, we first extract the multi-head attention weights of the global graph attention $\tA$ within the decoding layers. 
Afterwards, we apply a beam search decoding and a sentence-level aggregation to achieve $\tA'$. 
For the reference metric $\tR$, we compute a text similarity score between the generated sentences and the input paragraphs by calculating the ROUGE score. 
After obtaining $\tR$, we can calculate the correlation between $\tR$ and $\tA'$. This procedure is illustrated in \autoref{fig:rq2-pipeline}.

The specific measures taken are described in Section~\ref{sec:experimental:measures}.
Due to resource limitations, we perform our analysis on a reduced test dataset of $500$ generated summaries for MultiNews as well as WikiSum.
However, in pre-experiments we did not observe a difference when using only a fraction of this number of summaries.

All models are trained on three P100 GPUs with $16$GB memory each. The parallel training results in a speed up of the training procedure by increasing the effective batch size. This can also be interpreted as a gradient accumulation over three steps, when combining the three parallel processes.
Preliminary results indicate that both models converge within $100$ epochs. 
Therefore, we limit training to $100$ epochs with no early stopping.

\subsection{Hyperparameter Optimization}
\label{sec:experimental:hyperparameters}
\paragraph{Summarization using sentences vs. paragraphs.}
\label{sec:experimental:hyperparameters:sentence-paragraph}
As stated in \autoref{sec:research-methods}, the input data of the GraphSum model is first truncated and then padded to a fixed number of textual units as well as a fixed number of tokens within each textual unit.  
Just increasing the amount of textual units when training on sentence-level will result in a noticeable increase in memory consumption. 
Therefore, it is important to find a good balance between number of textual units and token per textual unit. 
For paragraph-level representation, GraphSum~\cite{li2020leveraging} used $30$ paragraphs with a maximum of $60$ tokens, resulting in a sequence of $1,800$ tokens for each set of multi-documents. 
For reasons of fair comparison, we chose the parameters for the sentence-level representation such that we have similar complexity.
We use $60$ sentences and with a maximum of $30$ tokens. 

Even though the GraphSum model operates on $1,800$ tokens for the sentence-level representation as well as for paragraph-level representation, the memory consumption for training on sentence-level is higher.
This behaviour arises from a larger number of vertices in the similarity graph, which also needs to fit into the GPU memory. 
Therefore, the batchsize when training on sentence-level has to be reduced by $25\%$ in comparison to paragraph-level.
Apart from this, we used the same values for the hyperparameters as \citeauthor{li2020leveraging} for training the model (\eg learning rate, beam size, optimizer, among others), which resulted in the best model performance.  

\paragraph{Source origin information via attention weights.}
As introduced in Section~\ref{sec:research-methods:orign}, an aggregation metric is applied to transform the \ac{AWD} containing token-level information to sentence-level. 
For this purpose we used the mean function.

\subsection{Measures}
\label{sec:experimental:measures}
For sentence-level vs. paragraph-level comparison, we use the ROUGE metric to evaluate the quality of the generated summaries against a reference summary. 
We consider ROUGE-1, which considers overlaps between uni-grams, ROUGE-2, which considers overlaps between bi-grams, and ROUGE-L, which considers the longest common sub sequence~\cite{lin2004rouge}.

For our source origin analysis, we calculate the ROUGE scores between the input paragraphs and each sentence in the generated summary.
Those ROUGE scores calculate the text similarity between the generated sentence and the input paragraphs. 
To this end, we split each input paragraph into its sentences and separately calculate the ROUGE score for each sentence. 
Those ROUGE scores are then averaged to obtain a paragraph-level $\tR$ score. 
This score indicates the source origin information of the generated sentence based on the text similarity to the input paragraphs.
After obtaining the source origin metric $\tR$, we calculate Pearson's correlation coefficient between the aggregated attention weights $\tA'$ and $\tR$, for each decoding layer and multi-head to evaluate how effectively attention weights in a transformer-based decoder can indicate source origin information.

\section{Results}
\label{sec:results}
\subsection{Summaries Using Sentences vs. Paragraphs}
\label{sec:results:RQ1}

\autoref{tab:rq1-results} shows the results for the comparison of sentence-level vs. paragraph-level summaries generated by our \ac{MDS} models on the test dataset.
With a batchsize of $3,000$ target tokens, the paragraph-level model outperforms the sentence-level model with regard to the ROUGE score by a small margin.
Furthermore, a larger batchsize of $4,000$ resulted in higher ROUGE scores for the paragraph-level models. 
These scores are similar to the best results reported by \citeauthor{li2020leveraging}, which confirms the validity of our results.

\begin{table}[h]
    \centering
        \begin{tabular}{|l|c|c|}
            \hline
            \thead{Textual \\ Unit} & \thead{ROUGE-F\\(1/2/L)}   & \thead{Batch Size \\ (number of target tokens)} \\
            \hline
            Sentences    & $43.82$/$15.85$/$40.31$ & $3,000$     \\
            \hline
            Paragraphs    & $44.82$/$16.89$/$41.17$ & $3,000$
            \\
            \hline
            Paragraphs    & $45.06$/$16.84$/$41.35$ & $4,000$     \\
            \hline
            \makecell{\citet{li2020leveraging}} & $45.71$/$17.12$/$41.99$ &  $4,096$ \\ \hline      
            \end{tabular}        
        \caption{Mean test results over 3 runs (ROUGE-F) for different model configurations trained on the MultiNews dataset.}
        \label{tab:rq1-results}
\end{table}

\subsection{Source Origin Analysis}
\label{sec:results:RQ2}

\begin{figure}[h]
    \captionsetup[subfigure]{textfont=normal}
    \centering
    \begin{subfigure}{0.49\columnwidth}
        \includegraphics[scale=0.35]{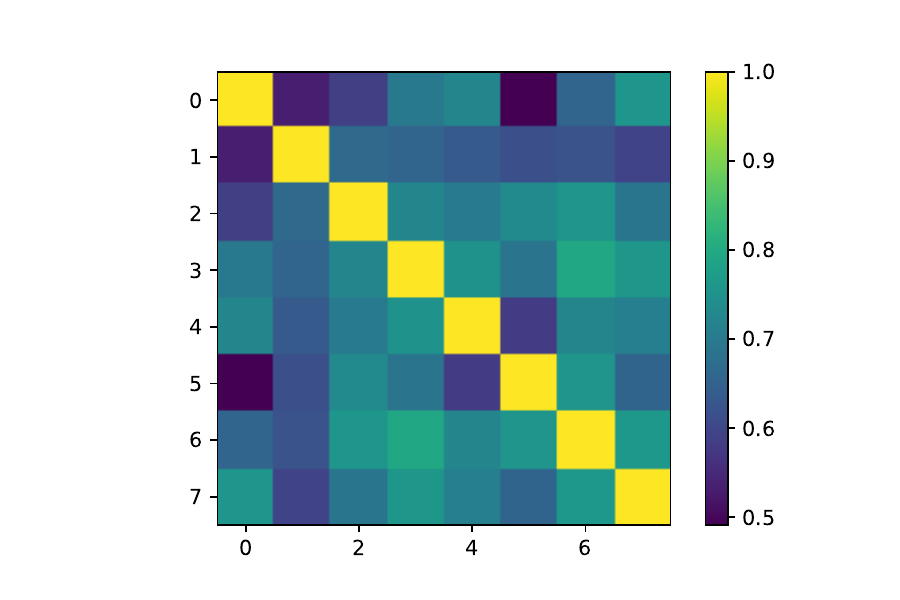}
        \caption{\textbf{MultiNews} Multi-heads}
        \label{fig:multi_head_corr}

    \end{subfigure}
    \begin{subfigure}{0.49\columnwidth}
\includegraphics[scale=0.35]{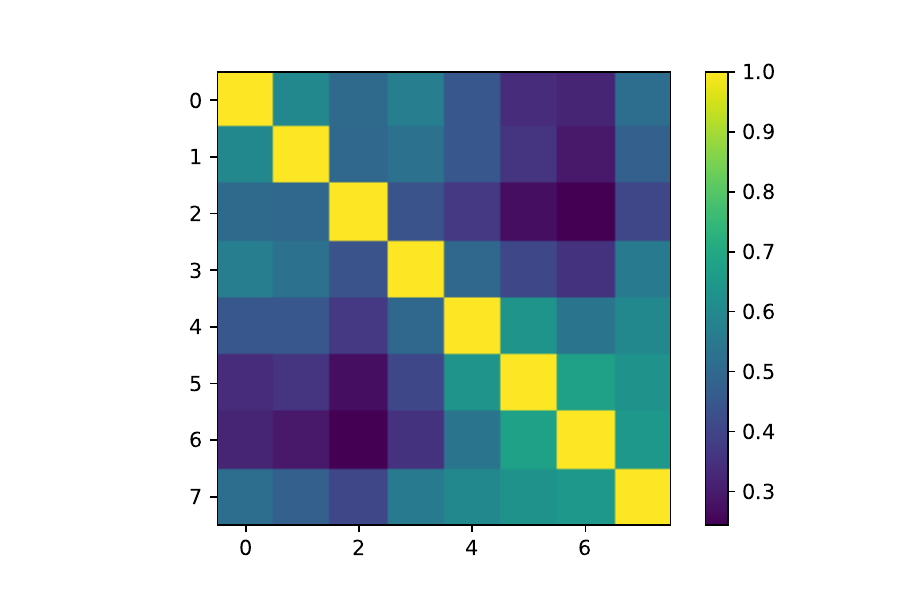}
        \caption{\textbf{MultiNews} Decoding layer}
        \label{fig:decoding_layer_corr}
    \end{subfigure}
    \\
    \begin{subfigure}{0.49\columnwidth}
    \includegraphics[scale=0.35]{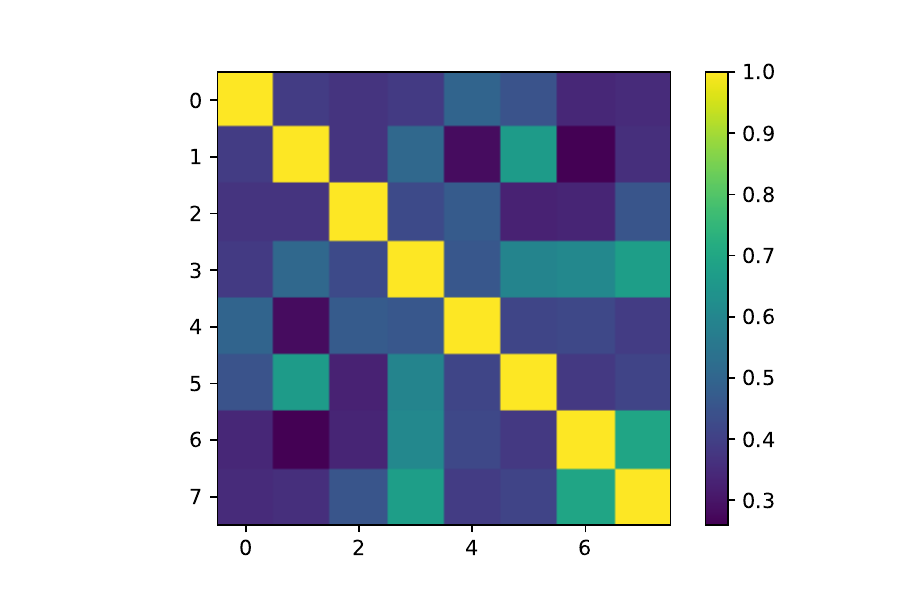}
    \caption{\textbf{WikiSum} Multi-heads}
    \label{fig:wikisum_multi_head_corr}
    \end{subfigure}
    \begin{subfigure}{0.49\columnwidth}
    \includegraphics[scale=0.35]{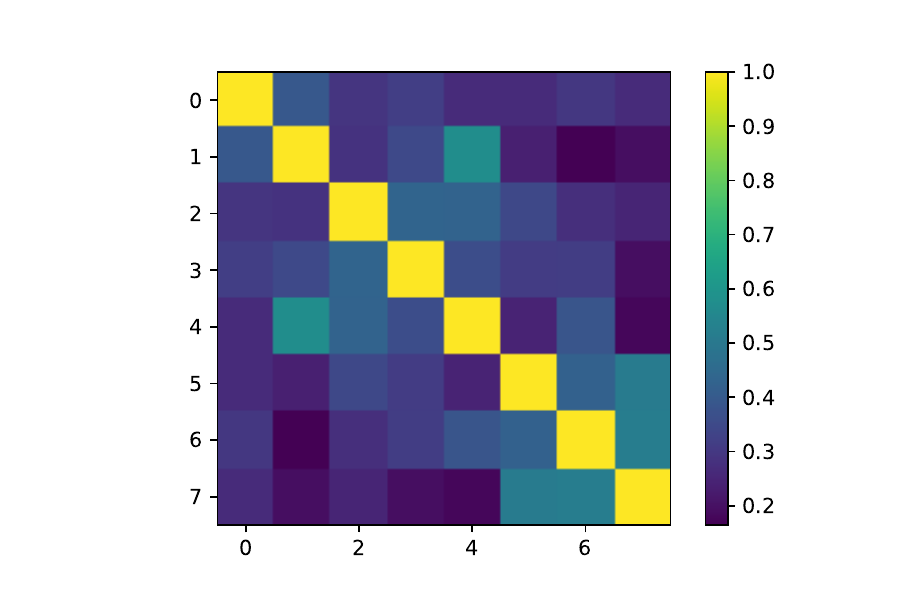}
    \caption{\textbf{WikiSum} Decoding layer}
    \label{fig:wikisum_decoding_layer_corr}
    \end{subfigure}

    \caption{Correlation between the attention weights of different multi-heads (left) and decoding layers (right), evaluated on the MultiNews and WikiSum dataset based on the global attention weights. The attention weights are extracted from a reduced test set. 
    }
    \label{fig:rq2-corr}
\end{figure}

The GraphSum model contains 8 decoding layers, while having 8 multi-heads for each attention mechanism. 
Thus, we analyze if the different multi-heads and decoding-layers attend different input paragraphs for the generation process. 
Our results show that the different multi-heads have a high correlation between each other with regard to their attention weights. 
This was especially present in the MultiNews dataset, for which this pattern is shown in \autoref{fig:multi_head_corr}. 
For the MultiNews dataset, the average correlation coefficient is $0.72$, while the minimum correlation coefficient between two multi-heads is $0.49$. 
For the WikiSum dataset, shown in \autoref{fig:wikisum_multi_head_corr}, the average coefficient is $0.51$, while the minimum correlation is $0.27$. 

Based on this result, we aggregated the attention weights of different multi-heads within each decoding layer for further analysis with regard to the correlation between the attention weights $\tA'$ and ROUGE-scores $\tR$ (see Section~\ref{sec:research-methods:orign}).
The correlation coefficient between different decoding layers varies. 
For the MultiNews dataset, shown in Figure~\ref{fig:decoding_layer_corr}, the average correlation coefficient is $0.53$, while the minimum correlation coefficient is $0.24$. 
For the WikiSum dataset (see Figure~\ref{fig:wikisum_decoding_layer_corr}) the average correlation is $0.41$, while the minimum correlation coefficient is $0.16$.
As shown in Figure~\ref{fig:decoding_layer_corr}, the correlation between subsequent layers in the MultiNews dataset is the highest.
This pattern is not as pronounced in the WikiSum dataset.

\begin{table}[h]
    \centering
        \begin{tabular}{|c|c|c|}
        \hline
            \multirow{2}{*}{\thead{Decoding layer}}  & \multicolumn{2}{c|}{\thead{Correlation($\tA'$, $\tR$)}}  \\ \cline{2-3}
             & \thead{MultiNews} & \thead{WikiSum}   \\ \hline
            $1$ & $0.18$/$0.20$/$0.19$ & $0.21$/$0.20$/$0.21$\\\hline
            $2$ & $0.16$/$0.18$/$0.17$ & \textit{0.02}/\textit{0.02}/\textit{0.02}\\\hline
            $3$ & \textit{0.13}/\textit{0.15}/\textit{0.14} & $0.22$/$0.23$/$0.23$\\\hline
            $4$ & $0.32$/$0.36$/$0.34$ & $0.10$/$0.12$/$0.11$ \\\hline
            $5$ & $0.46$/$0.53$/$0.50$ & $0.15$/$0.19$/$0.17$ \\\hline
            $6$ & \textbf{0.56}/\textbf{0.69}/\textbf{0.63} & $0.41$/$0.52$/$0.46$ \\\hline
            $7$ & $0.56$/$0.69$/$0.63$  & $0.46$/$0.57$/$0.52$ \\\hline
            $8$ & $0.47$/$0.54$/$0.51$ & \textbf{0.49/0.58/0.53}\\\hline
        \end{tabular}        
        \caption{Correlation for the different decoding layers for the MultiNews dataset and WikiSum dataset, based on the global attention weights ($\tA'$) with regard to text-similarity information obtained by ROUGE-scores ($\tR$), which contains information about ROUGE-F(1/2/L) information.  
        Highest scores are shown in bold, lowest in italics.
        The attention weights and ROUGE-scores are extracted from a reduced test set.}
        \label{tab:rq2}
\end{table}

The correlation between the aggregated \ac{AWD} of each decoding layer and the ROUGE-scores $\tR$ are shown in \autoref{tab:rq2}. 
These results show that the decoding layers form two clusters. 
The first cluster contains the first layers, which average a lower correlation coefficient compared to the later layers, which make up the second cluster. 

\begin{figure*}
\captionsetup[subfigure]{textfont=normal}
    \centering
    \subcaptionbox{\textbf{MultiNews} Decoding Layer 3}
        {\includegraphics[width=0.99\columnwidth]{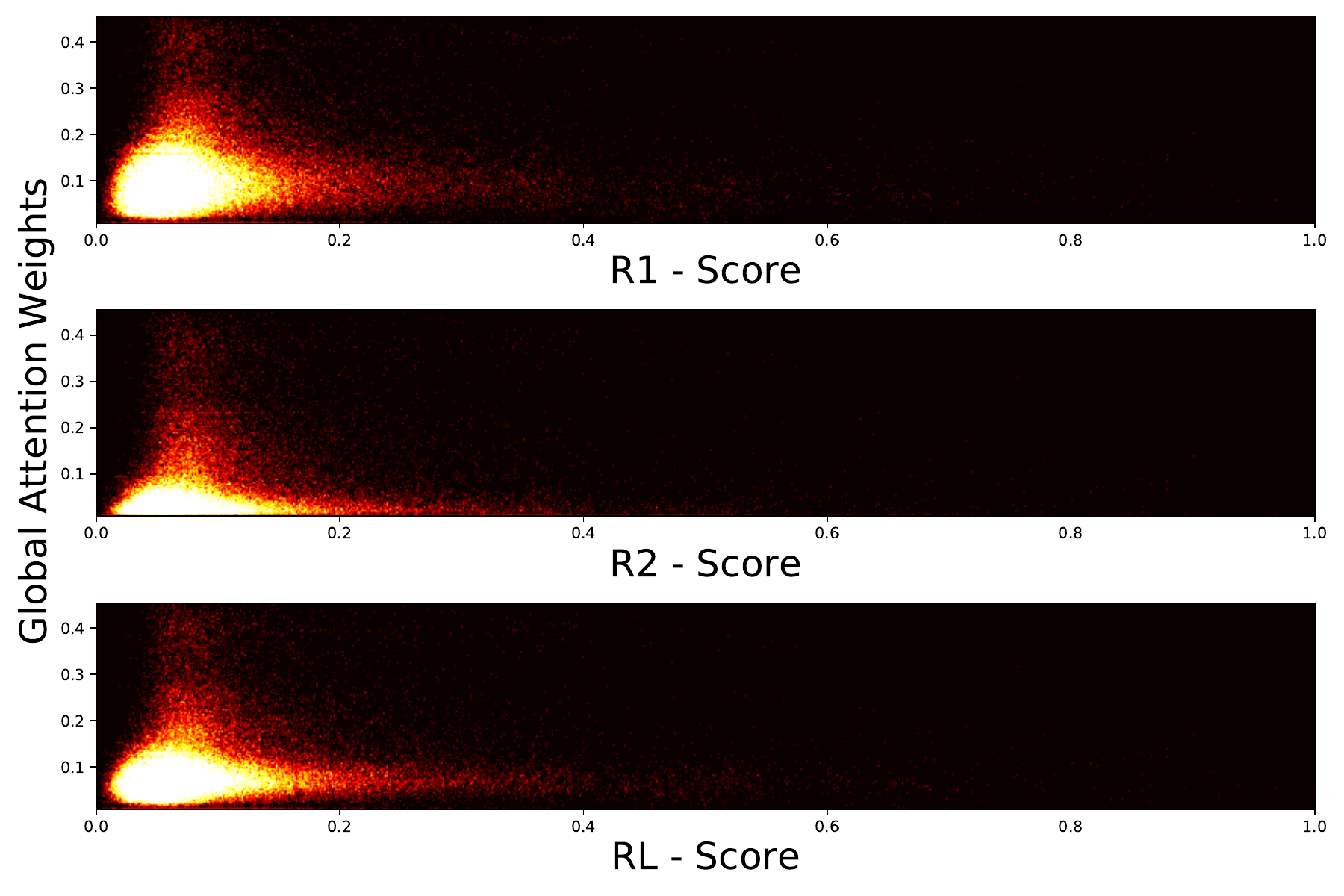}}
    \subcaptionbox{\textbf{MultiNews} Decoding Layer 6}
        {\includegraphics[width=0.99\columnwidth]{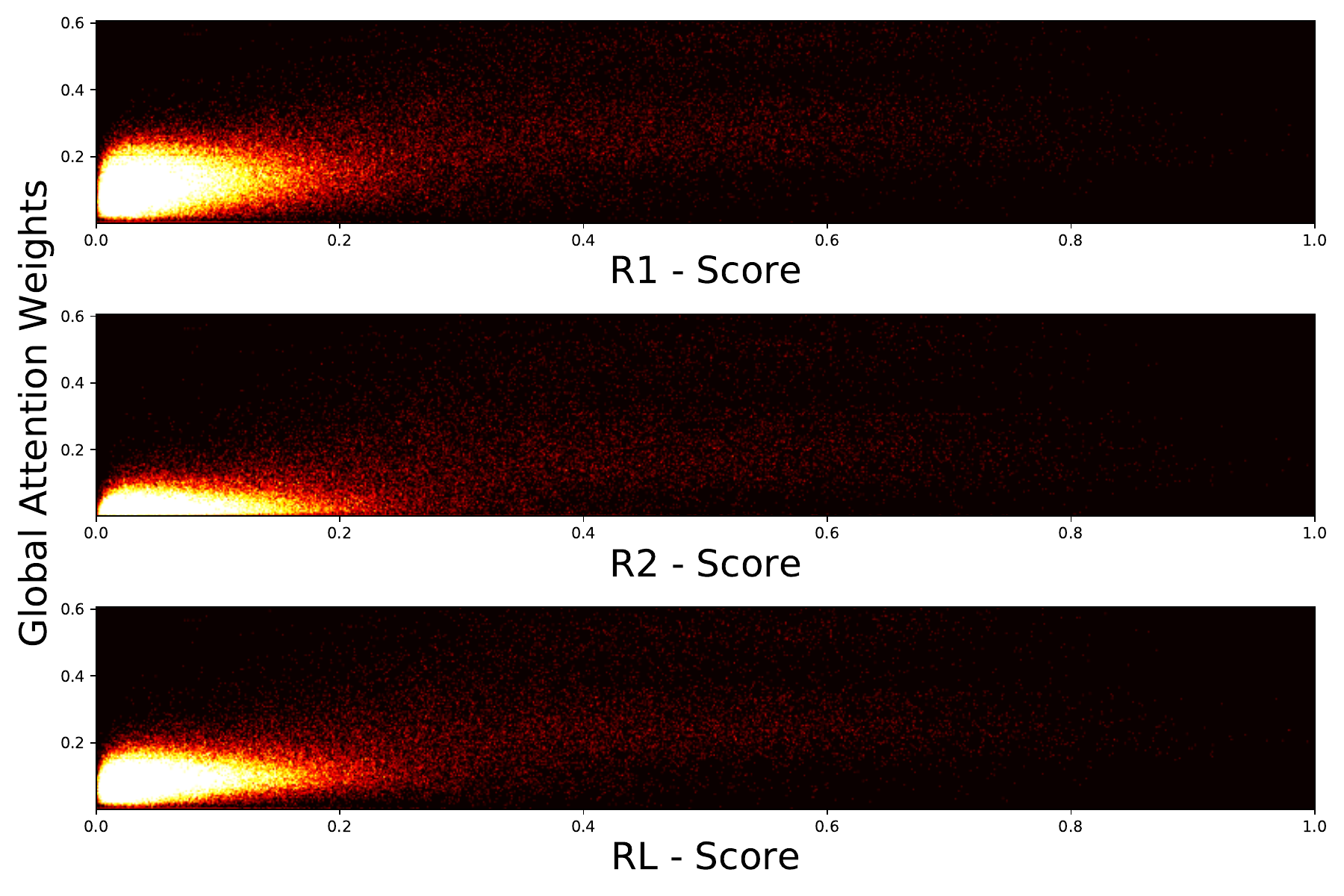}}

    \caption{Visualization of correlation between ROUGE scores indicating source origin information via text similarity and the aggregated global attention weights for the decoding layer with the lowest correlation (left) and the decoding layer with the highest correlation (right) of the MultiNews dataset. A reduced test set is used for evaluation.  
    }
    \label{fig:rq2-multinews-correlation}
\end{figure*}

\begin{figure*}
\captionsetup[subfigure]{textfont=normal}
    \centering
    \subcaptionbox{\textbf{WikiSum} Decoding Layer 2}
        {\includegraphics[width=0.99\columnwidth]{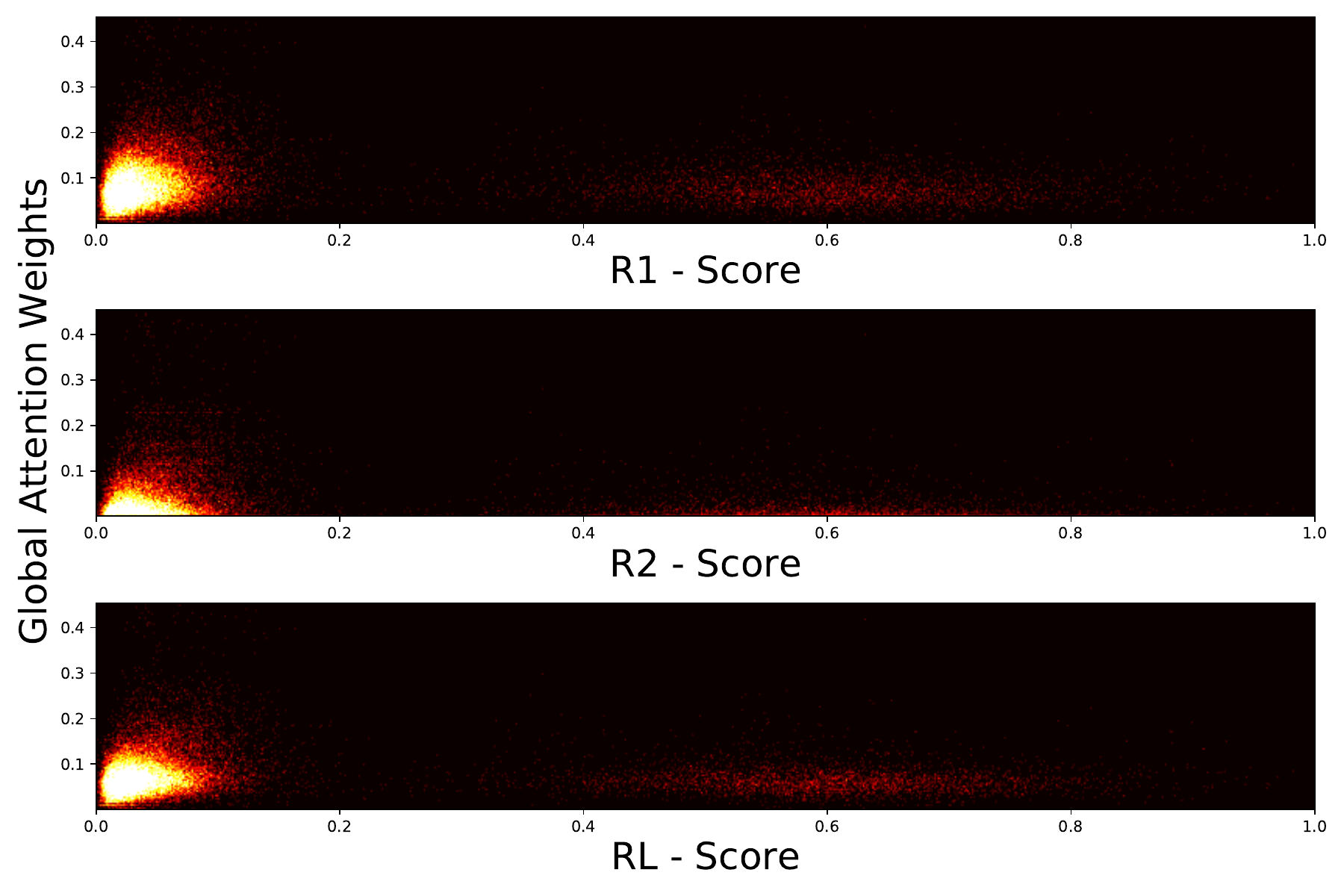}}
    \subcaptionbox{\textbf{WikiSum} Decoding Layer 8}
        {\includegraphics[width=0.99\columnwidth]{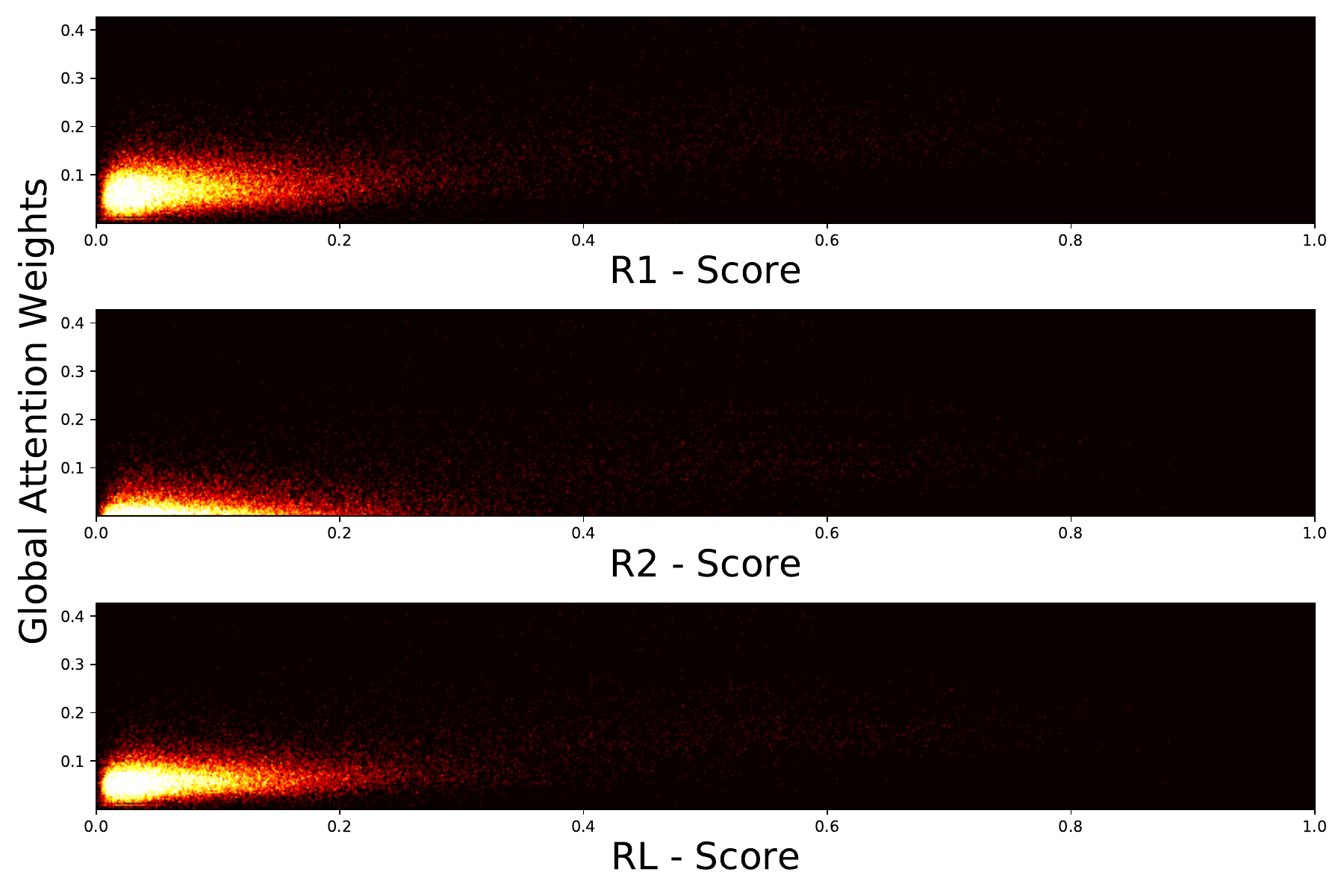}}
        \caption{Visualization of correlation between ROUGE scores indicating source origin information via text similarity and the aggregated global attention weights for the decoding layer with the lowest correlation (left) and the decoding layer with the highest correlation (right) of the WikiSum dataset. A reduced test set is used for evaluation.  
    }
    \label{fig:rq2-wiki-correlation}
\end{figure*}

For each cluster, we consider a single layer for further analysis, namely the layer with the lowest correlation (italics) within the first cluster and the layer with the highest correlation (bold font) within the second cluster. 
The correlation between $\tA'$ with their respective $\tR$ is visualized in \autoref{fig:rq2-multinews-correlation} for the MultiNews dataset and in \autoref{fig:rq2-wiki-correlation} for the WikiSum dataset. 
\extended{The correlation for further decoding layers can be found in \autoref{sec:appendix:correlation_visualization}.}

Overall, we observe a high correlation coefficient between the attention weights and reference metric for the source origin analysis.  
This holds particularly true for the deeper decoding layers as shown in Table~\ref{tab:rq2}.
Thus, we investigate if the generated summaries of GraphSum follow the pattern of positional bias. 
For this purpose, we can use the attention weights to retrieve the input paragraph, which provides the most information for a generated sentence. 
To analyze if the positional bias is present in the generated summaries, we aggregate this information over all generated summaries. 
Since the input of the GraphSum \cite{li2020leveraging} model is interpreted as a sequence of tokens, we need information about which paragraphs correspond to which input document. 
The WikiSum dataset does not contain this information, therefore our analysis of positional bias is limited to the MultiNews dataset.

\begin{figure}[h]
    \centering
    \subcaptionbox{Decoding Layer 3}
        {\includegraphics[width=0.99\columnwidth]{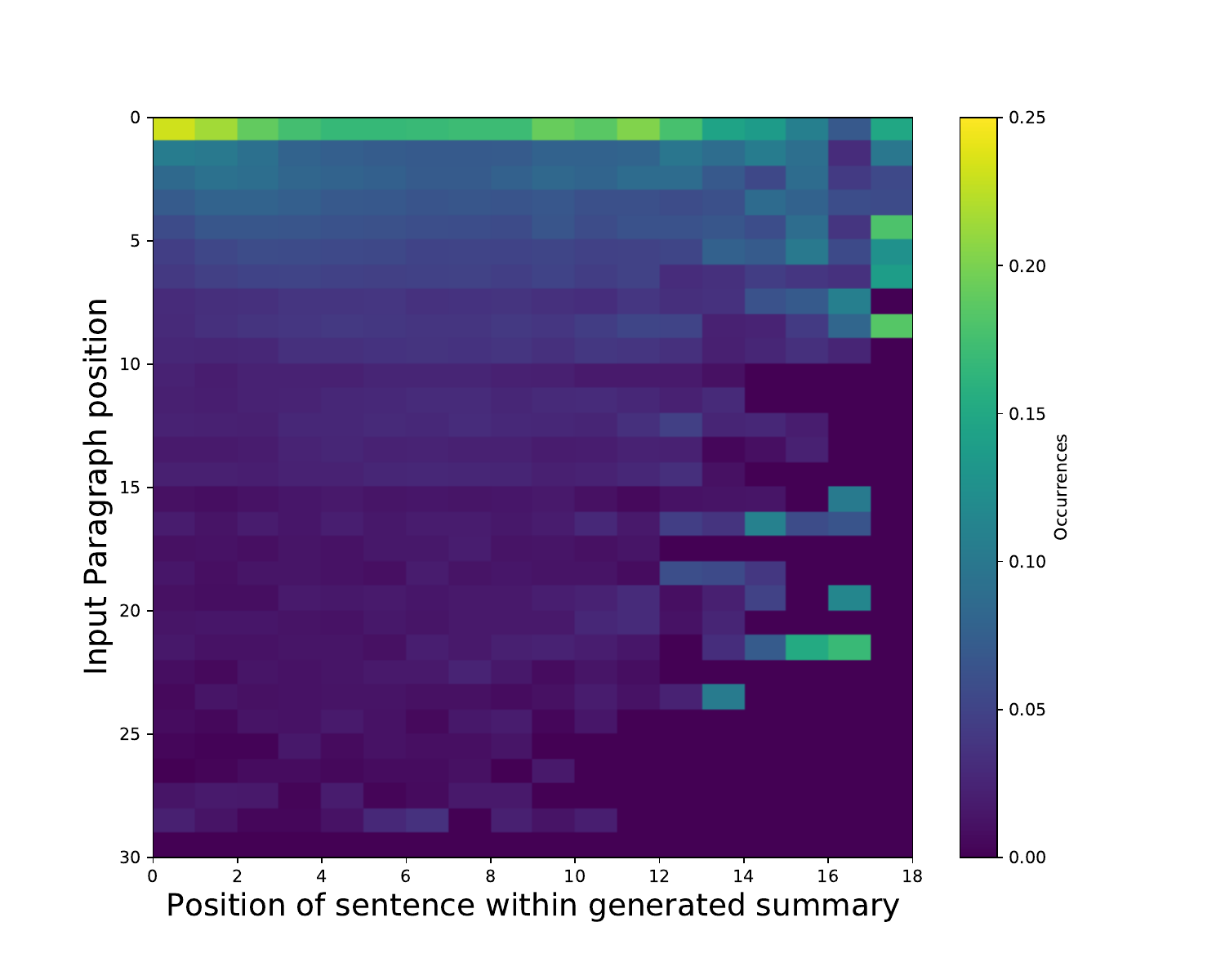}}
    \subcaptionbox{Decoding Layer 6}
        {\includegraphics[width=0.99\columnwidth]{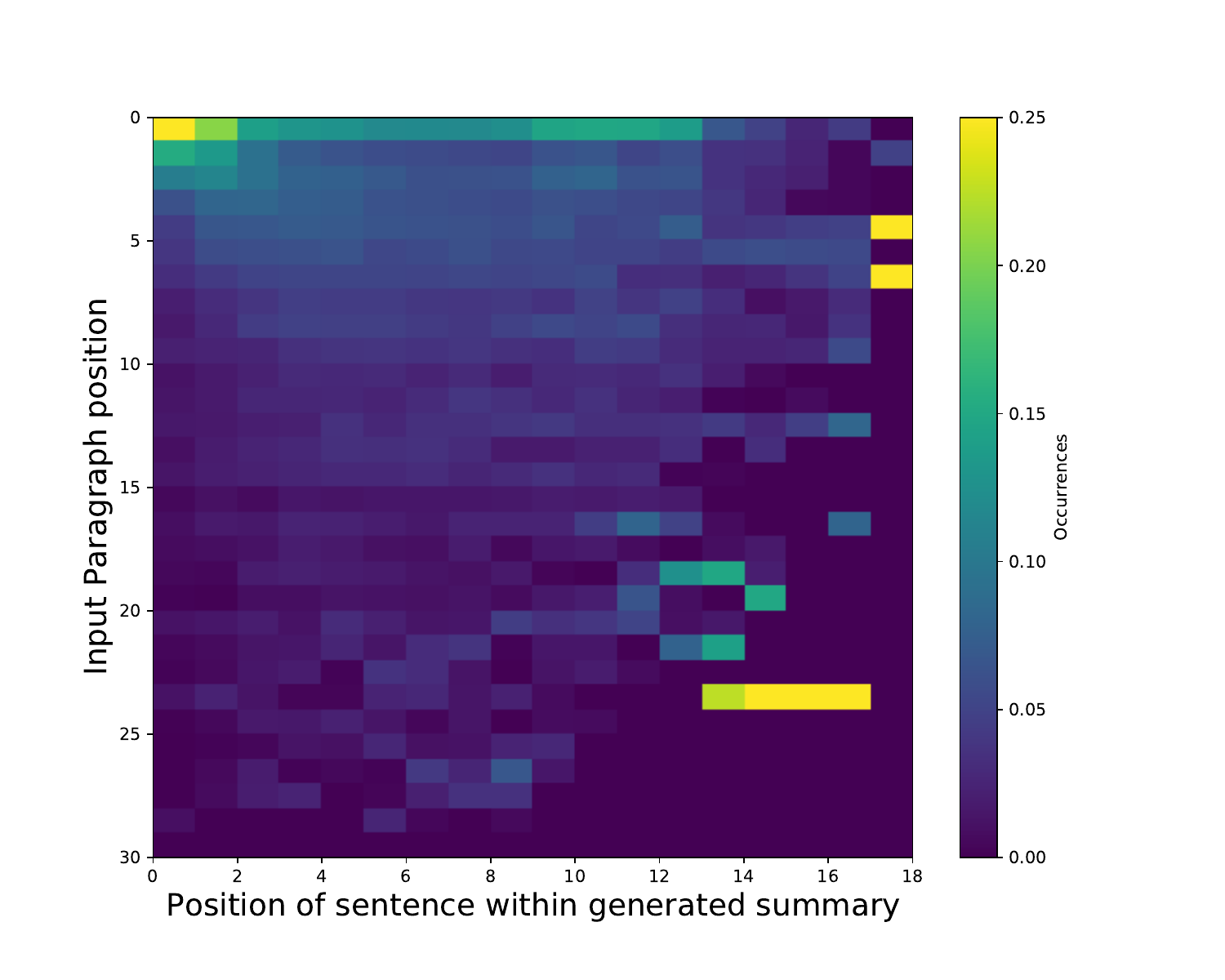}}

    \caption{Heatmap to visualize the paragraph positions, which were attended the most for each generated sentence. The analysis is performed on the MutliNews dataset.}
    \label{fig:rq2-pos-bias-att}
\end{figure}

According to the attention weights, for most generated sentences of the summary, the first paragraphs of each input document are attended the most.
This pattern can be seen in the heat-maps shown in \autoref{fig:rq2-pos-bias-att} for layer 3, with the lowest correlation, and layer 6, with the highest correlation.
The heat-maps visualize where the information about which paragraph was attended the most for a generated sentence according to the attention weights. 
Note that the generated summaries have different numbers of sentences. 
This results in a bias as there are fewer summaries with a high number of sentences in the test set.
This behaviour can especially be observed in \autoref{fig:rq2-pos-bias-att} (b), where artefacts can be seen for later sentences of generated summaries with a high number of sentences \ie in the right columns of the heatmap.
\extended{The plots for further decoding layers can be found in \autoref{sec:appendix:positional_bias}.}

\section{Discussion}
\label{sec:discussion}
\subsection{Summaries Using Sentences vs. Paragraphs}

\begin{figure}[h]
    \includegraphics[width=\columnwidth]{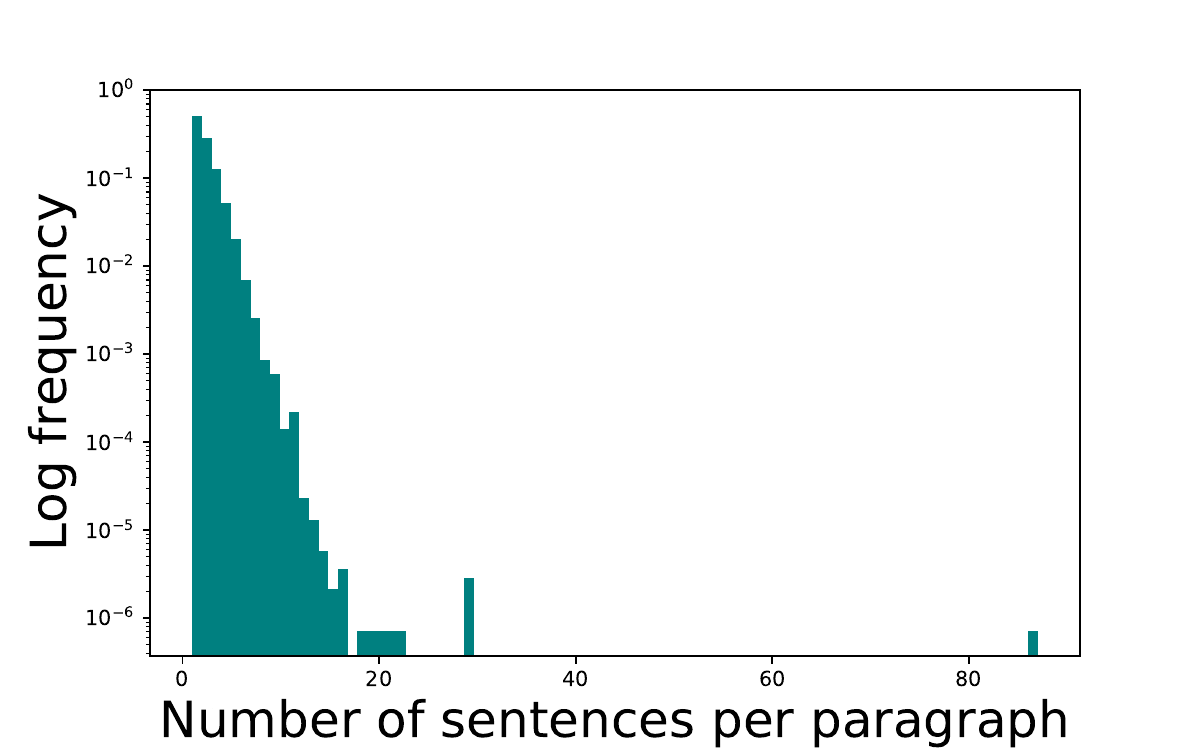}
    \centering
    \caption{
    Number of sentences per paragraph for the MultiNews dataset. Histogram is density normed and has a log-scaled y-axis to illustrate outliers.}
    \label{fig:rq1-paragraph-lengths}
\end{figure}

Our experiments indicate that paragraph-level models perform better than sentence-level models on the MultiNews dataset. 
However, the margin is rather small.
We have analyzed the MultiNews dataset in more detail.
We observe that the paragraphs often contain a single sentence as shown by the long-tail distribution in \autoref{fig:rq1-paragraph-lengths}. 
The y-axis of the right plot is log-scaled in order to visualize the outliers.
On average, a paragraph contains only $1.84$ sentences. 
The vast majority of the paragraphs contain less then three sentences.
However, there are also outliers in the sense that the paragraphs contain a lot of sentences.

It appears that paragraphs are mainly used as structural aid in news articles to improve readability and not to separate topics, resulting in a low number of sentences per paragraph. 
This may explain why a sentence-level representation and the paragraph-level representation perform quite similar.
We would expect similar observations in other \ac{MDS} domains where paragraphs are rather short, too.

As the number of sentences in the MultiNews dataset exceeds the number of paragraphs by a factor of $1.84$, more textual units have to be considered when working with sentence-level representations to cover the same information.
To keep the complexity of the model similar to paragraph-level representations, the number of tokens per textual unit had to be decreased. 
This has to be done as the MultiNews dataset contains outlier news articles  with a lot of sentences (see \autoref{fig:rq1-paragraph-lengths}). 
This can lead to an issue that paragraphs, which only contain a single sentence, are truncated and therefore possibly important information is lost. Due to the heterogeneity of the MultiNews dataset with regard to usage of paragraphs, truncating and padding with a fixed number of tokens will always lead to a trade-off between a lower batch size and an information loss. Both aspects can lead to a decreased model performance. 
Due to resource limitations, we adjusted the padding and truncation parameters in order to achieve a batch size of $3,000$, for which the paragraph-level model achieved good results.
Furthermore, sentence-level representation results in more vertices in the graph structure which leads to a higher memory consumption. %
As our results showed, an increased batchsize improves model performance on paragraph level. 
With limited resources, it is expected that a paragraph-level model outperforms a sentence-level model as it is possible to utilize larger batchsizes.

We were not able to train a sentence-level model on a batchsize of $4,000$, which produced the best results for paragraph-level. 
It might be that a sentence-level model could outperform the paragraph-level model for a higher batchsize. 
However, preliminary results with different batchsizes showed that the paragraph-level model outperformed the sentence-level model for all configurations. 
As the memory consumption depends on the padding procedure as well as the batchsize, an increase of available resources also enables the possibility to increase the number of tokens per textual unit, which could counteract the issue of losing information by truncation.

While for our comparisons we only constructed a similarity graph based on TF-IDF scores, graphs capturing lexical relations as well as discourse relations~\cite{christensen-etal-2013-towards} are also shown to be beneficial for text summarization tasks~\cite{li2020leveraging}. 
However, the results of~\citeauthor{li2020leveraging} were similar for all different graph representations.
Thus, we omitted them here in our study.
In addition, \citeauthor{christensen-etal-2013-towards}'s discourse graph generally operates on token-level and captures the discourse relations between sentences and between not paragraphs~\cite{christensen-etal-2013-towards}. 
Therefore, we did not use this graph type for our comparisons.

\subsection{Source origin analysis}

Our results on the source origin analysis using the attention weights show that the multi-heads in a single decoding layer attend mostly the same paragraphs.
This is indicated by a high correlation between the attention weights. 
We argue that the multi-heads are still able to extract different information while attending the same paragraphs, since the output of a multi-head attention mechanism depends on trainable parameters, which extract information based on the attended input paragraphs. 
Our experiments show that the \ac{AWD} can be used to indicate which paragraphs provided relevant information for the generated summaries, especially the attention weights of deeper layers are suitable for this task. 
This was expected because the output of the transformer layer is a linear transformed weighted average of the input paragraphs as shown in \autoref{eq:attention}. 
Furthermore, in deep neural networks for computer vision, it is known that early layers learn to extract high level features while layers closer to the output extract more task specific features~\cite{yosinski2014transfer}. 
Similar effects may apply to language models, as indicated by the success of transfer-learning and fine-tuning of BERT for language processing tasks~\cite{devlin2018bert}. 

Even though we expected a gradually change for the correlation of consecutive layers, it is particularly noticeable that the decoding layers can be clustered into two sets with regard to the similarity of their attention weights. 
There is a sudden jump of the correlations observed from layer 4 to 5 for the MultiNews dataset and 5 to 6 for the WikiSum dataset, see Table~\ref{tab:rq2}.

Based on the analysis of GraphSum's attention mechanisms, our results show that the graph-based \ac{MDS} model generates summaries mostly from the first  paragraphs of each input document in the MultiNews dataset.
Therefore, we assume that the GraphSum model indeed learns the positional bias of the MultiNews dataset.
However, it is not clear whether the model attends the first paragraphs based on their positional encoding or their actual content. 
Therefore it might be interesting to investigate how well a trained \ac{MDS} model can perform on shuffled input documents, where the order of textual units (\ie sentences, paragraphs or documents) is changed.
This would allow to understand, if a trained model extracts information based on the position of a textual unit or its content.
We leave this as part of future work.

We assumed that text similarity based on the ROUGE score can be used as a reference metric in our source origin analysis. 
One may argue that the ROUGE metric might be unsuitable as reference metric for abstractive summaries, which are heavily based on synonyms and/or paraphrasing, as the ROUGE metric indicates source origin information by calculating text overlaps between generated summaries and input paragraphs. 
However, abstractive summaries should still contain immutable key words (\eg names, dates, locations) which can be captured by the ROUGE metric. 
Therefore, simple paraphrasing or the usage of synonyms for unimportant information should not distort the effectiveness of the reference metric.

An even better reference metric could be achieved if experts annotated the MultiNews dataset with source origin information manually.
This labeled dataset could then be used to further evaluate the possibility of using attention weights to extract source origin information.
Furthermore, to aggregate the token-level information of the attention weights to sentence-level, we calculated the mean value of the attention weights of the tokens. 
We assume that the mean is suitable for this task under the assumption that the model generates coherent summaries and therefore subsequent tokens should have a similar source origin. 
Other aggregation metrics like median could be evaluated in future work.
Furthermore, the transformer architecture of the GraphSum model also calculates a local graph attention which attends input tokens. 
This local graph attention also directly influences the generated tokens. 
A combination of both attention weight distributions, \ie global and local, could lead to even better result in terms of source origin information.
In general, we assume that the attention weights of any transformer based architecture in the \ac{MDS} domain should be suitable for extracting source origin information. 
To verify this assumption, our approach could be applied to other transformer-based models.

\section{Conclusion}
\label{sec:conclusion}

Out experiments show that the summaries generated from a sentence-level representation of the MultiNews dataset using an extended GraphSum model are of similar quality than the paragraph-level representation.
However, the sentence-level representation requires more computational resources as more vertices and edges are need in the graph-based model.
Thus, subsequently we investigated the source origin information by analyzing the attention weights on the paragraph-level representation.
Here, we revealed that the attention weights of later decoding layers of the GraphSum's decoder architecture contain valuable information about which input paragraphs provide the most information for the generated summaries.
These weights in the MultiNews dataset correlate with the presence of a positional bias, \ie that the most important information is located in the first sentences. Therefore, we conclude that the attention weights successfully learn a positional bias.
This sheds some lights on better explaining how abstract \ac{MDS} using graph-based transformer models work.
Note, the analysis of positional bias could not be done on the WikiSum dataset as it does not contain information about which paragraphs correspond to which input document. 
In future work, it will be interesting to investigate this property in the WikiSum dataset and other domains, as well as to apply our analyses to other types of transformer-based models. 

\paragraph{\dataavailabilityandreproducibility}
The code of this work is available on GitHub: \url{https://github.com/arnelochner/GBTBMDS}.
 
\paragraph{\acknowledgments}
\creditmasterproject{WS 2020/2021}

\extended{\bibliographystyle{ACM-Reference-Format}}

\bibliography{references-base}


\begin{thebibliography}{22}


\ifx \showCODEN    \undefined \def \showCODEN     #1{\unskip}     \fi
\ifx \showDOI      \undefined \def \showDOI       #1{#1}\fi
\ifx \showISBNx    \undefined \def \showISBNx     #1{\unskip}     \fi
\ifx \showISBNxiii \undefined \def \showISBNxiii  #1{\unskip}     \fi
\ifx \showISSN     \undefined \def \showISSN      #1{\unskip}     \fi
\ifx \showLCCN     \undefined \def \showLCCN      #1{\unskip}     \fi
\ifx \shownote     \undefined \def \shownote      #1{#1}          \fi
\ifx \showarticletitle \undefined \def \showarticletitle #1{#1}   \fi
\ifx \showURL      \undefined \def \showURL       {\relax}        \fi
\providecommand\bibfield[2]{#2}
\providecommand\bibinfo[2]{#2}
\providecommand\natexlab[1]{#1}
\providecommand\showeprint[2][]{arXiv:#2}

\bibitem[\protect\citeauthoryear{Bahdanau, Cho, and Bengio}{Bahdanau
  et~al\mbox{.}}{2015}]%
        {bahdanau15attention}
\bibfield{author}{\bibinfo{person}{Dzmitry Bahdanau},
  \bibinfo{person}{Kyunghyun Cho}, {and} \bibinfo{person}{Yoshua Bengio}.}
  \bibinfo{year}{2015}\natexlab{}.
\newblock \showarticletitle{Neural Machine Translation by Jointly Learning to
  Align and Translate}. In \bibinfo{booktitle}{\emph{3rd International
  Conference on Learning Representations, {ICLR} 2015, San Diego, CA, USA, May
  7-9, 2015, Conference Track Proceedings}},
  \bibfield{editor}{\bibinfo{person}{Yoshua Bengio} {and} \bibinfo{person}{Yann
  LeCun}} (Eds.).
\newblock
\urldef\tempurl%
\url{http://arxiv.org/abs/1409.0473}
\showURL{%
\tempurl}


\bibitem[\protect\citeauthoryear{Christensen, {Mausam}, Soderland, and
  Etzioni}{Christensen et~al\mbox{.}}{2013}]%
        {christensen-etal-2013-towards}
\bibfield{author}{\bibinfo{person}{Janara Christensen},
  \bibinfo{person}{{Mausam}}, \bibinfo{person}{Stephen Soderland}, {and}
  \bibinfo{person}{Oren Etzioni}.} \bibinfo{year}{2013}\natexlab{}.
\newblock \showarticletitle{Towards Coherent Multi-Document Summarization}. In
  \bibinfo{booktitle}{\emph{Proceedings of the 2013 Conference of the North
  {A}merican Chapter of the Association for Computational Linguistics: Human
  Language Technologies}}. \bibinfo{publisher}{Association for Computational
  Linguistics}, \bibinfo{address}{Atlanta, Georgia},
  \bibinfo{pages}{1163--1173}.
\newblock
\urldef\tempurl%
\url{https://www.aclweb.org/anthology/N13-1136}
\showURL{%
\tempurl}


\bibitem[\protect\citeauthoryear{Devlin, Chang, Lee, and Toutanova}{Devlin
  et~al\mbox{.}}{2018}]%
        {devlin2018bert}
\bibfield{author}{\bibinfo{person}{Jacob Devlin}, \bibinfo{person}{Ming-Wei
  Chang}, \bibinfo{person}{Kenton Lee}, {and} \bibinfo{person}{Kristina
  Toutanova}.} \bibinfo{year}{2018}\natexlab{}.
\newblock \showarticletitle{Bert: Pre-training of deep bidirectional
  transformers for language understanding}.
\newblock \bibinfo{journal}{\emph{arXiv preprint arXiv:1810.04805}}
  (\bibinfo{year}{2018}).
\newblock


\bibitem[\protect\citeauthoryear{Erkan and Radev}{Erkan and Radev}{2004}]%
        {erkan2004lexrank}
\bibfield{author}{\bibinfo{person}{G{\"u}nes Erkan} {and}
  \bibinfo{person}{Dragomir~R Radev}.} \bibinfo{year}{2004}\natexlab{}.
\newblock \showarticletitle{Lexrank: Graph-based lexical centrality as salience
  in text summarization}.
\newblock \bibinfo{journal}{\emph{Journal of artificial intelligence research}}
   \bibinfo{volume}{22} (\bibinfo{year}{2004}), \bibinfo{pages}{457--479}.
\newblock


\bibitem[\protect\citeauthoryear{Fabbri, Li, She, Li, and Radev}{Fabbri
  et~al\mbox{.}}{2019}]%
        {fabbri2019multinews}
\bibfield{author}{\bibinfo{person}{Alexander~R. Fabbri}, \bibinfo{person}{Irene
  Li}, \bibinfo{person}{Tianwei She}, \bibinfo{person}{Suyi Li}, {and}
  \bibinfo{person}{Dragomir~R. Radev}.} \bibinfo{year}{2019}\natexlab{}.
\newblock \bibinfo{title}{Multi-News: a Large-Scale Multi-Document
  Summarization Dataset and Abstractive Hierarchical Model}.
\newblock
\newblock
\showeprint[arxiv]{1906.01749}~[cs.CL]


\bibitem[\protect\citeauthoryear{Huang, Wu, and Wang}{Huang
  et~al\mbox{.}}{2020}]%
        {huang2020knowledge}
\bibfield{author}{\bibinfo{person}{Luyang Huang}, \bibinfo{person}{Lingfei Wu},
  {and} \bibinfo{person}{Lu Wang}.} \bibinfo{year}{2020}\natexlab{}.
\newblock \bibinfo{title}{Knowledge Graph-Augmented Abstractive Summarization
  with Semantic-Driven Cloze Reward}.
\newblock
\newblock
\showeprint[arxiv]{2005.01159}~[cs.CL]


\bibitem[\protect\citeauthoryear{Jung, Kang, Mentch, and Hovy}{Jung
  et~al\mbox{.}}{2019}]%
        {jung2019earlier}
\bibfield{author}{\bibinfo{person}{Taehee Jung}, \bibinfo{person}{Dongyeop
  Kang}, \bibinfo{person}{Lucas Mentch}, {and} \bibinfo{person}{Eduard Hovy}.}
  \bibinfo{year}{2019}\natexlab{}.
\newblock \bibinfo{title}{Earlier Isn't Always Better: Sub-aspect Analysis on
  Corpus and System Biases in Summarization}.
\newblock
\newblock
\showeprint[arxiv]{1908.11723}~[cs.CL]


\bibitem[\protect\citeauthoryear{Kedzie, McKeown, and au2}{Kedzie
  et~al\mbox{.}}{2019}]%
        {kedzie2019content}
\bibfield{author}{\bibinfo{person}{Chris Kedzie}, \bibinfo{person}{Kathleen
  McKeown}, {and} \bibinfo{person}{Hal Daume~III au2}.}
  \bibinfo{year}{2019}\natexlab{}.
\newblock \bibinfo{title}{Content Selection in Deep Learning Models of
  Summarization}.
\newblock
\newblock
\showeprint[arxiv]{1810.12343}~[cs.CL]


\bibitem[\protect\citeauthoryear{Kipf and Welling}{Kipf and Welling}{2016}]%
        {kipf2016semi}
\bibfield{author}{\bibinfo{person}{Thomas~N Kipf} {and} \bibinfo{person}{Max
  Welling}.} \bibinfo{year}{2016}\natexlab{}.
\newblock \showarticletitle{Semi-supervised classification with graph
  convolutional networks}.
\newblock \bibinfo{journal}{\emph{arXiv preprint arXiv:1609.02907}}
  (\bibinfo{year}{2016}).
\newblock


\bibitem[\protect\citeauthoryear{Kudo and Richardson}{Kudo and
  Richardson}{2018a}]%
        {kudo2018sentencepiece}
\bibfield{author}{\bibinfo{person}{Taku Kudo} {and} \bibinfo{person}{John
  Richardson}.} \bibinfo{year}{2018}\natexlab{a}.
\newblock \showarticletitle{Sentencepiece: A simple and language independent
  subword tokenizer and detokenizer for neural text processing}.
\newblock \bibinfo{journal}{\emph{arXiv preprint arXiv:1808.06226}}
  (\bibinfo{year}{2018}).
\newblock


\bibitem[\protect\citeauthoryear{Kudo and Richardson}{Kudo and
  Richardson}{2018b}]%
        {kudo-richardson-2018-sentencepiece}
\bibfield{author}{\bibinfo{person}{Taku Kudo} {and} \bibinfo{person}{John
  Richardson}.} \bibinfo{year}{2018}\natexlab{b}.
\newblock \showarticletitle{{S}entence{P}iece: A simple and language
  independent subword tokenizer and detokenizer for Neural Text Processing}. In
  \bibinfo{booktitle}{\emph{Proceedings of the 2018 Conference on Empirical
  Methods in Natural Language Processing: System Demonstrations}}.
  \bibinfo{publisher}{Association for Computational Linguistics},
  \bibinfo{address}{Brussels, Belgium}, \bibinfo{pages}{66--71}.
\newblock
\urldef\tempurl%
\url{https://doi.org/10.18653/v1/D18-2012}
\showDOI{\tempurl}


\bibitem[\protect\citeauthoryear{Li, Xiao, Liu, Wu, Wang, and Du}{Li
  et~al\mbox{.}}{2020}]%
        {li2020leveraging}
\bibfield{author}{\bibinfo{person}{Wei Li}, \bibinfo{person}{Xinyan Xiao},
  \bibinfo{person}{Jiachen Liu}, \bibinfo{person}{Hua Wu},
  \bibinfo{person}{Haifeng Wang}, {and} \bibinfo{person}{Junping Du}.}
  \bibinfo{year}{2020}\natexlab{}.
\newblock \bibinfo{title}{Leveraging Graph to Improve Abstractive
  Multi-Document Summarization}.
\newblock
\newblock
\showeprint[arxiv]{2005.10043}~[cs.CL]


\bibitem[\protect\citeauthoryear{Lin}{Lin}{2004}]%
        {lin2004rouge}
\bibfield{author}{\bibinfo{person}{Chin-Yew Lin}.}
  \bibinfo{year}{2004}\natexlab{}.
\newblock \showarticletitle{{ROUGE}: A Package for Automatic Evaluation of
  Summaries}. In \bibinfo{booktitle}{\emph{Text Summarization Branches Out}}.
  \bibinfo{publisher}{Association for Computational Linguistics},
  \bibinfo{address}{Barcelona, Spain}, \bibinfo{pages}{74--81}.
\newblock
\urldef\tempurl%
\url{https://www.aclweb.org/anthology/W04-1013}
\showURL{%
\tempurl}


\bibitem[\protect\citeauthoryear{Lin and Hovy}{Lin and Hovy}{2002}]%
        {lin2002single}
\bibfield{author}{\bibinfo{person}{Chin-Yew Lin} {and} \bibinfo{person}{Eduard
  Hovy}.} \bibinfo{year}{2002}\natexlab{}.
\newblock \showarticletitle{From single to multi-document summarization}. In
  \bibinfo{booktitle}{\emph{Proceedings of the 40th annual meeting of the
  association for computational linguistics}}. \bibinfo{pages}{457--464}.
\newblock


\bibitem[\protect\citeauthoryear{Liu, Saleh, Pot, Goodrich, Sepassi, Kaiser,
  and Shazeer}{Liu et~al\mbox{.}}{2018}]%
        {liu2018generating}
\bibfield{author}{\bibinfo{person}{Peter~J. Liu}, \bibinfo{person}{Mohammad
  Saleh}, \bibinfo{person}{Etienne Pot}, \bibinfo{person}{Ben Goodrich},
  \bibinfo{person}{Ryan Sepassi}, \bibinfo{person}{Lukasz Kaiser}, {and}
  \bibinfo{person}{Noam Shazeer}.} \bibinfo{year}{2018}\natexlab{}.
\newblock \bibinfo{title}{Generating Wikipedia by Summarizing Long Sequences}.
\newblock
\newblock
\showeprint[arxiv]{1801.10198}~[cs.CL]


\bibitem[\protect\citeauthoryear{Liu and Lapata}{Liu and Lapata}{2019a}]%
        {liu2019hierarchical}
\bibfield{author}{\bibinfo{person}{Yang Liu} {and} \bibinfo{person}{Mirella
  Lapata}.} \bibinfo{year}{2019}\natexlab{a}.
\newblock \bibinfo{title}{Hierarchical Transformers for Multi-Document
  Summarization}.
\newblock
\newblock
\showeprint[arxiv]{1905.13164}~[cs.CL]


\bibitem[\protect\citeauthoryear{Liu and Lapata}{Liu and Lapata}{2019b}]%
        {liu2019text}
\bibfield{author}{\bibinfo{person}{Yang Liu} {and} \bibinfo{person}{Mirella
  Lapata}.} \bibinfo{year}{2019}\natexlab{b}.
\newblock \showarticletitle{Text summarization with pretrained encoders}.
\newblock \bibinfo{journal}{\emph{arXiv preprint arXiv:1908.08345}}
  (\bibinfo{year}{2019}).
\newblock


\bibitem[\protect\citeauthoryear{Vaswani, Shazeer, Parmar, Uszkoreit, Jones,
  Gomez, Kaiser, and Polosukhin}{Vaswani et~al\mbox{.}}{2017}]%
        {vaswani2017attention}
\bibfield{author}{\bibinfo{person}{Ashish Vaswani}, \bibinfo{person}{Noam
  Shazeer}, \bibinfo{person}{Niki Parmar}, \bibinfo{person}{Jakob Uszkoreit},
  \bibinfo{person}{Llion Jones}, \bibinfo{person}{Aidan~N. Gomez},
  \bibinfo{person}{Lukasz Kaiser}, {and} \bibinfo{person}{Illia Polosukhin}.}
  \bibinfo{year}{2017}\natexlab{}.
\newblock \bibinfo{title}{Attention Is All You Need}.
\newblock
\newblock
\showeprint[arxiv]{1706.03762}~[cs.CL]


\bibitem[\protect\citeauthoryear{Wang, Liu, Zheng, Qiu, and Huang}{Wang
  et~al\mbox{.}}{2020}]%
        {wang2020heterogeneous}
\bibfield{author}{\bibinfo{person}{Danqing Wang}, \bibinfo{person}{Pengfei
  Liu}, \bibinfo{person}{Yining Zheng}, \bibinfo{person}{Xipeng Qiu}, {and}
  \bibinfo{person}{Xuanjing Huang}.} \bibinfo{year}{2020}\natexlab{}.
\newblock \bibinfo{title}{Heterogeneous Graph Neural Networks for Extractive
  Document Summarization}.
\newblock
\newblock
\showeprint[arxiv]{2004.12393}~[cs.CL]


\bibitem[\protect\citeauthoryear{Xu, Gan, Cheng, and Liu}{Xu
  et~al\mbox{.}}{2020}]%
        {xu2020discourseaware}
\bibfield{author}{\bibinfo{person}{Jiacheng Xu}, \bibinfo{person}{Zhe Gan},
  \bibinfo{person}{Yu Cheng}, {and} \bibinfo{person}{Jingjing Liu}.}
  \bibinfo{year}{2020}\natexlab{}.
\newblock \bibinfo{title}{Discourse-Aware Neural Extractive Text
  Summarization}.
\newblock
\newblock
\showeprint[arxiv]{1910.14142}~[cs.CL]


\bibitem[\protect\citeauthoryear{Yasunaga, Zhang, Meelu, Pareek, Srinivasan,
  and Radev}{Yasunaga et~al\mbox{.}}{2017}]%
        {yasunaga-etal-2017-graph}
\bibfield{author}{\bibinfo{person}{Michihiro Yasunaga}, \bibinfo{person}{Rui
  Zhang}, \bibinfo{person}{Kshitijh Meelu}, \bibinfo{person}{Ayush Pareek},
  \bibinfo{person}{Krishnan Srinivasan}, {and} \bibinfo{person}{Dragomir
  Radev}.} \bibinfo{year}{2017}\natexlab{}.
\newblock \showarticletitle{Graph-based Neural Multi-Document Summarization}.
  In \bibinfo{booktitle}{\emph{Proceedings of the 21st Conference on
  Computational Natural Language Learning ({C}o{NLL} 2017)}}.
  \bibinfo{publisher}{Association for Computational Linguistics},
  \bibinfo{address}{Vancouver, Canada}, \bibinfo{pages}{452--462}.
\newblock
\urldef\tempurl%
\url{https://doi.org/10.18653/v1/K17-1045}
\showDOI{\tempurl}


\bibitem[\protect\citeauthoryear{Yosinski, Clune, Bengio, and Lipson}{Yosinski
  et~al\mbox{.}}{2014}]%
        {yosinski2014transfer}
\bibfield{author}{\bibinfo{person}{Jason Yosinski}, \bibinfo{person}{Jeff
  Clune}, \bibinfo{person}{Yoshua Bengio}, {and} \bibinfo{person}{Hod Lipson}.}
  \bibinfo{year}{2014}\natexlab{}.
\newblock \showarticletitle{How Transferable Are Features in Deep Neural
  Networks?}. In \bibinfo{booktitle}{\emph{Proceedings of the 27th
  International Conference on Neural Information Processing Systems - Volume
  2}} (Montreal, Canada) \emph{(\bibinfo{series}{NIPS'14})}.
  \bibinfo{publisher}{MIT Press}, \bibinfo{address}{Cambridge, MA, USA},
  \bibinfo{pages}{3320–3328}.
\newblock


\end{thebibliography}
\newpage

\appendix
\section*{Supplementary Materials}
\label{sec:appendix}

\section{Correlation Visualization}
\label{sec:appendix:correlation_visualization}

\begin{figure*}[p]
\captionsetup[subfigure]{textfont=normal}
    \centering
    \subcaptionbox{\textbf{MultiNews} Decoding Layer 1}
        {\includegraphics[width=0.9\columnwidth]{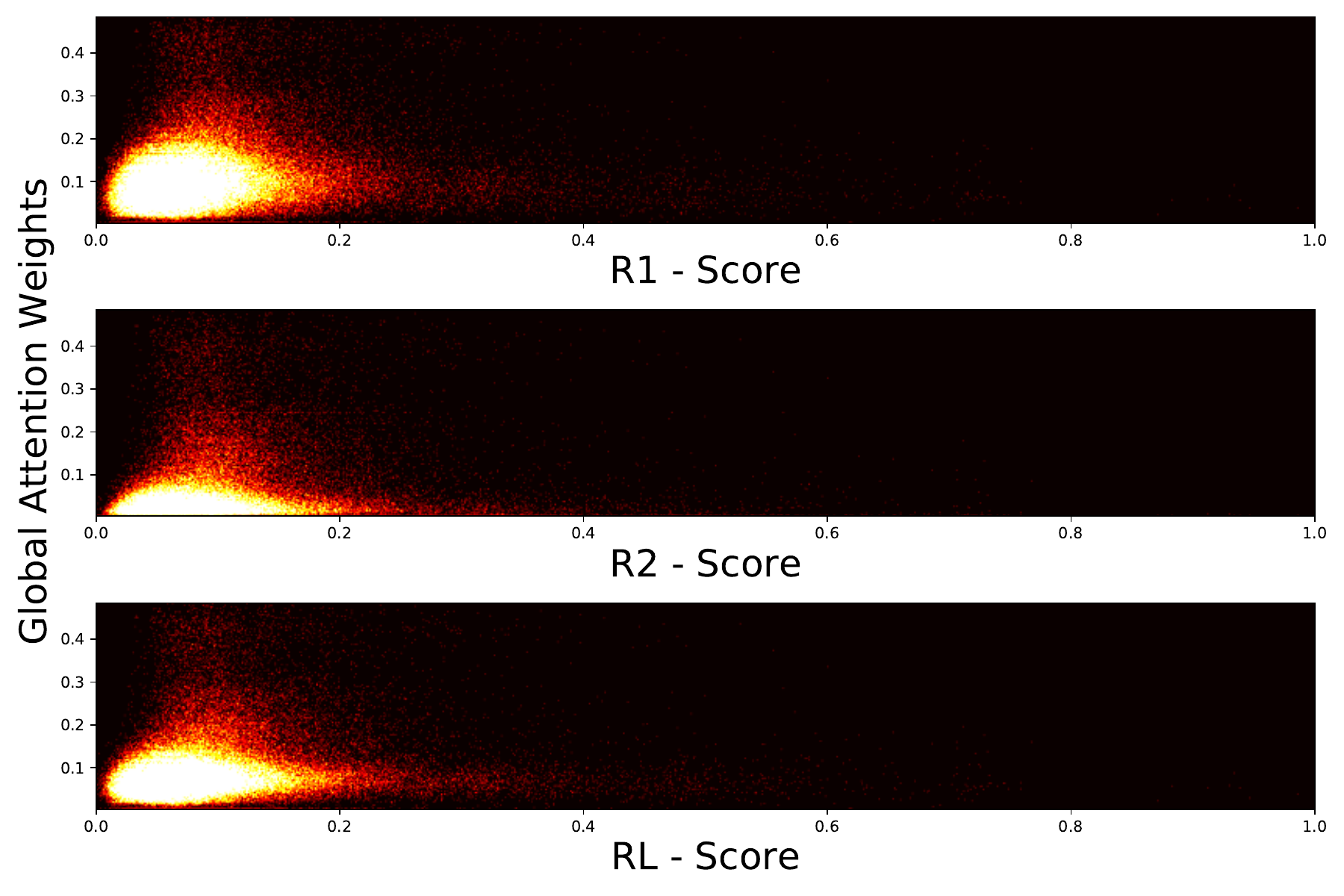}}
        \subcaptionbox{\textbf{MultiNews} Decoding Layer 2}
        {\includegraphics[width=0.9\columnwidth]{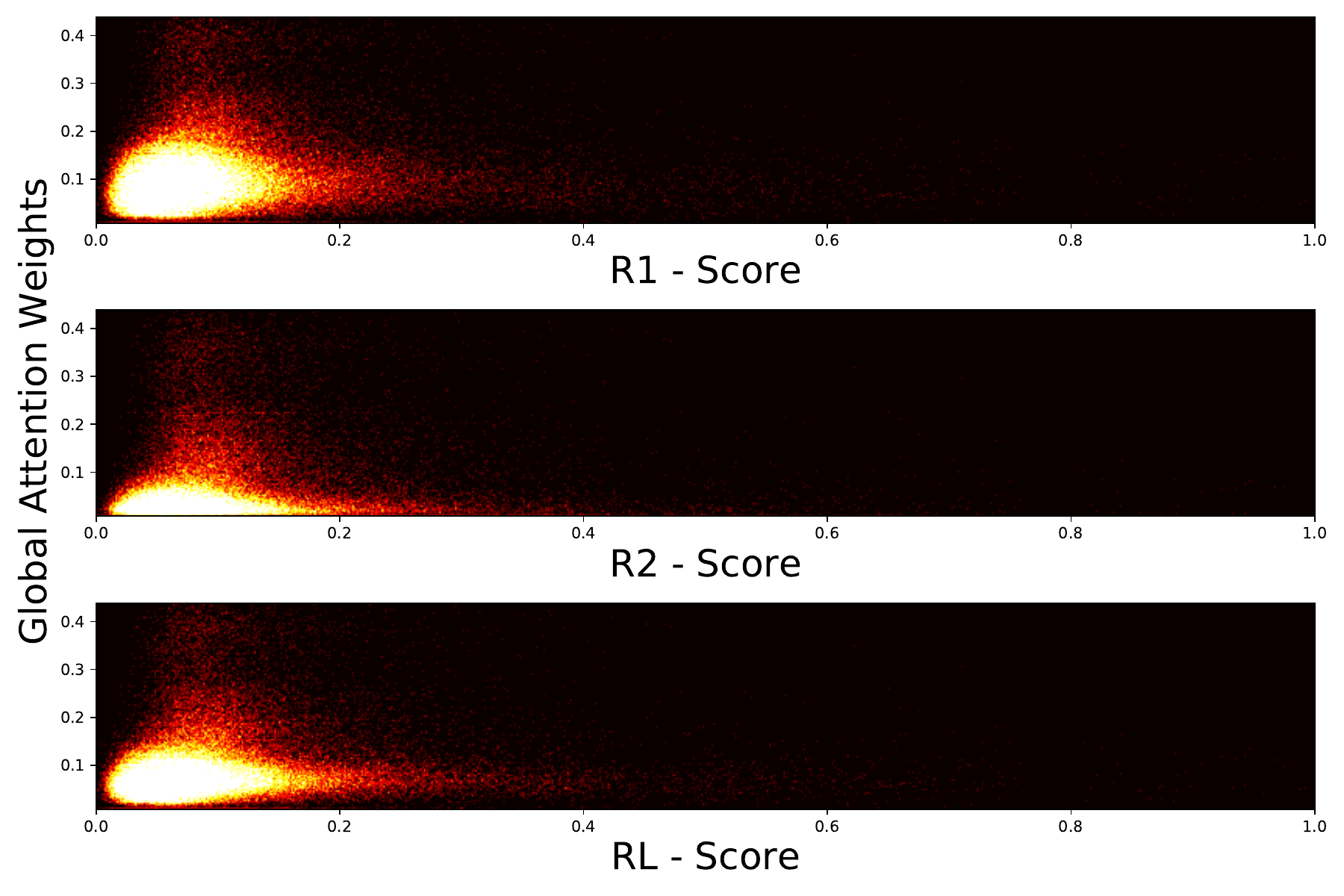}}
        \subcaptionbox{\textbf{MultiNews} Decoding Layer 3}
        {\includegraphics[width=0.9\columnwidth]{images/RQ2/multinews_correlation/correlation_plot_layer_3.pdf}}
        \subcaptionbox{\textbf{MultiNews} Decoding Layer 4}
        {\includegraphics[width=0.9\columnwidth]{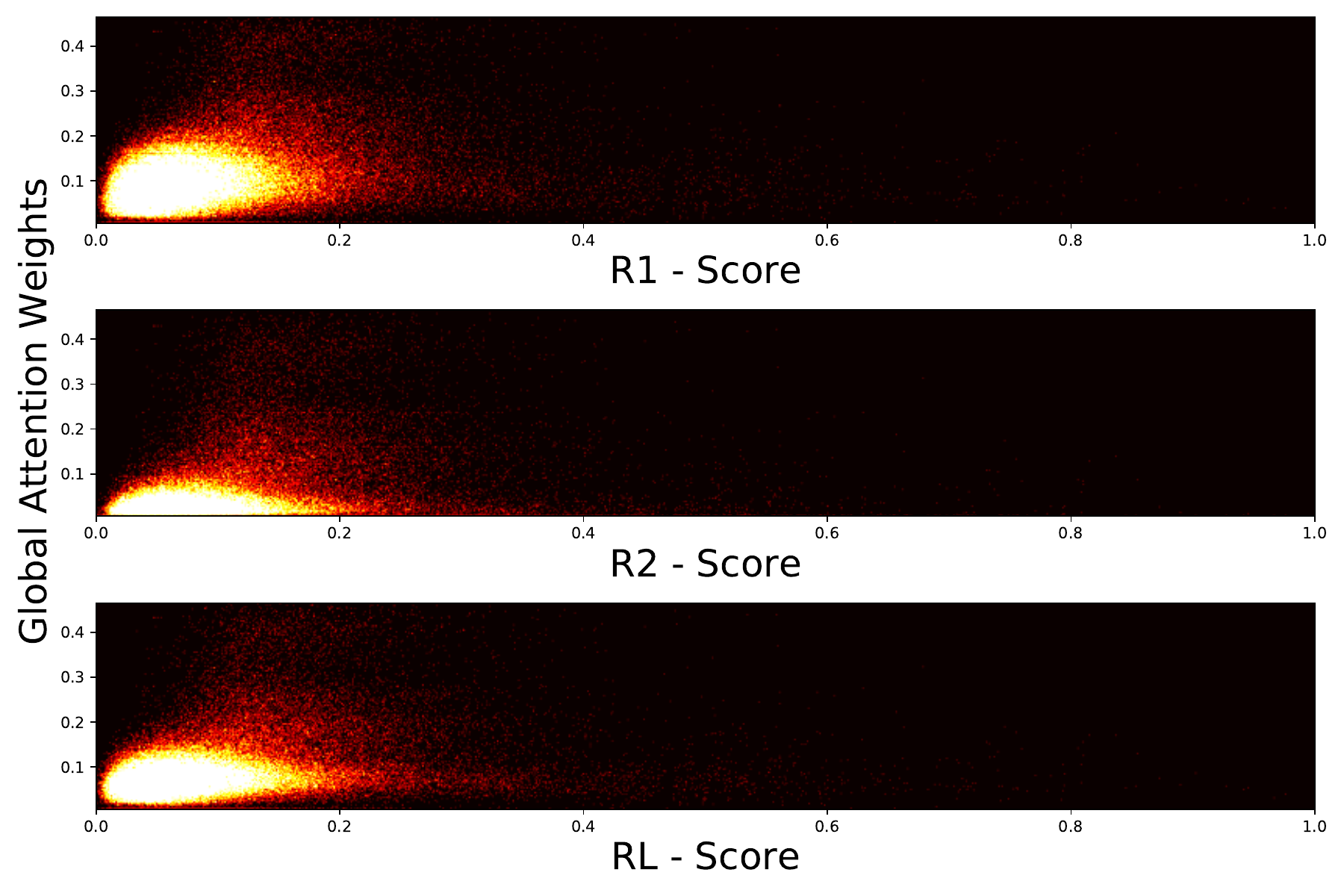}}
        \subcaptionbox{\textbf{MultiNews} Decoding Layer 5}
        {\includegraphics[width=0.9\columnwidth]{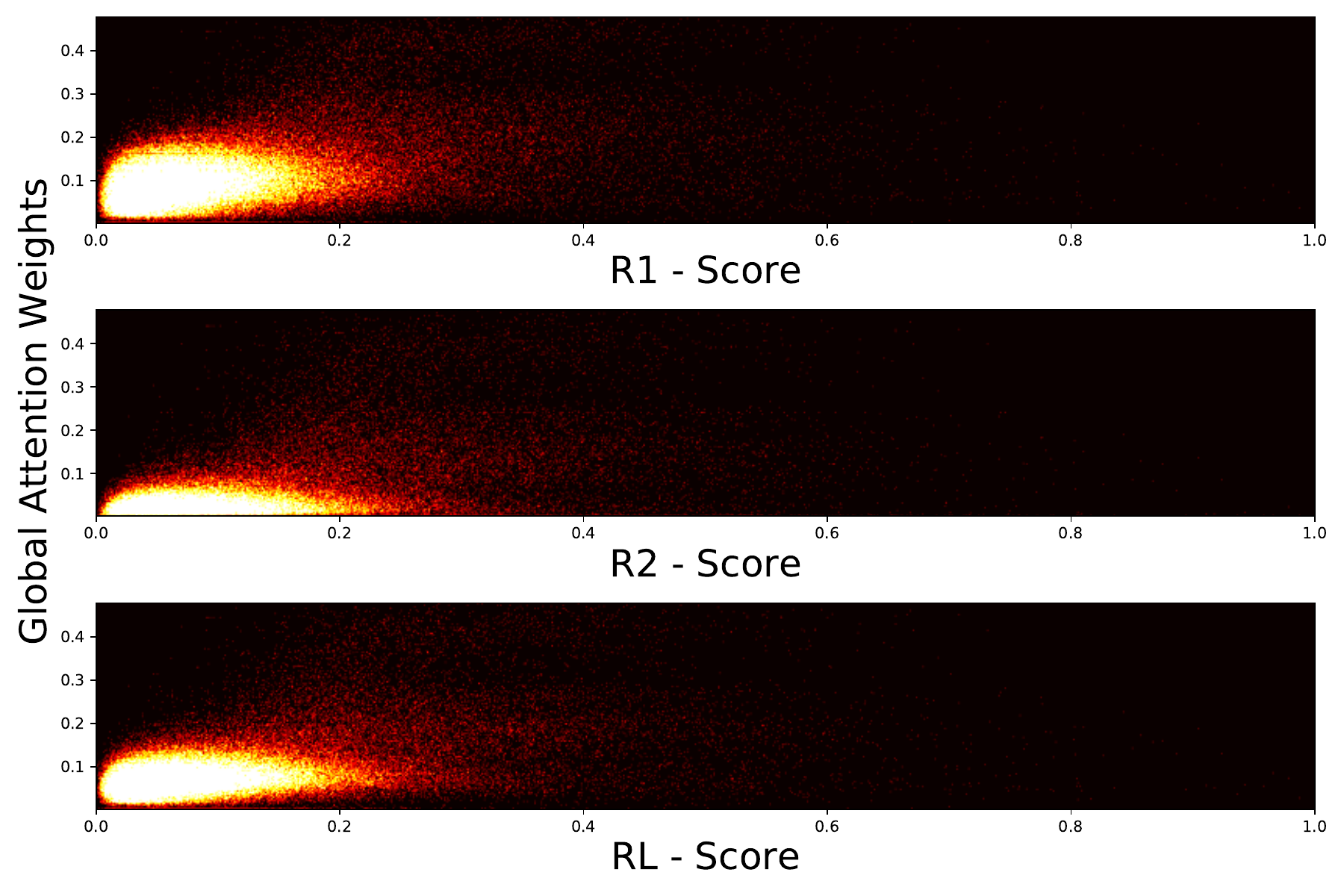}}
        \subcaptionbox{\textbf{MultiNews} Decoding Layer 6}
        {\includegraphics[width=0.9\columnwidth]{images/RQ2/multinews_correlation/correlation_plot_layer_5.pdf}}
        \subcaptionbox{\textbf{MultiNews} Decoding Layer 7}
        {\includegraphics[width=0.9\columnwidth]{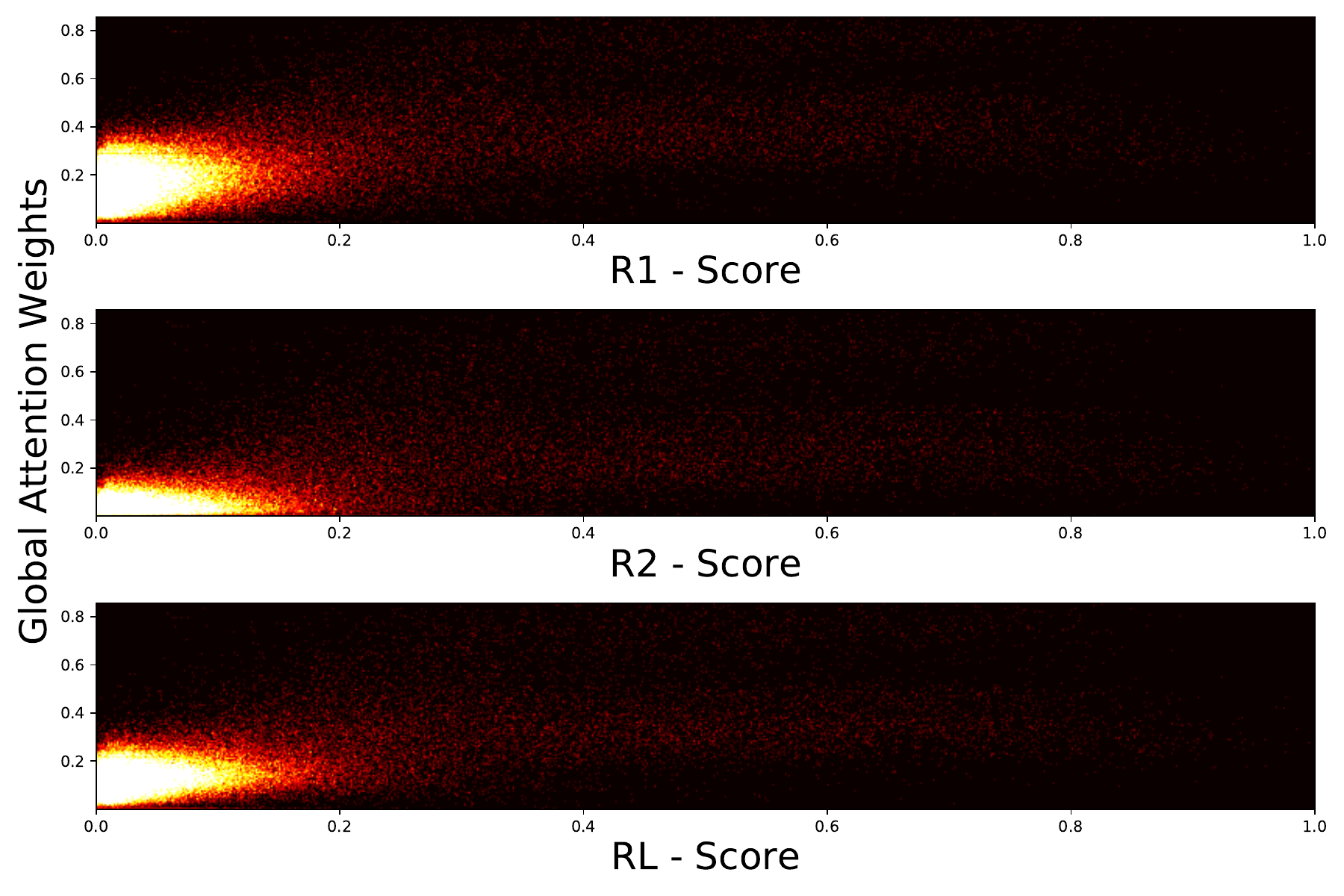}}
        \subcaptionbox{\textbf{MultiNews} Decoding Layer 8}
        {\includegraphics[width=0.9\columnwidth]{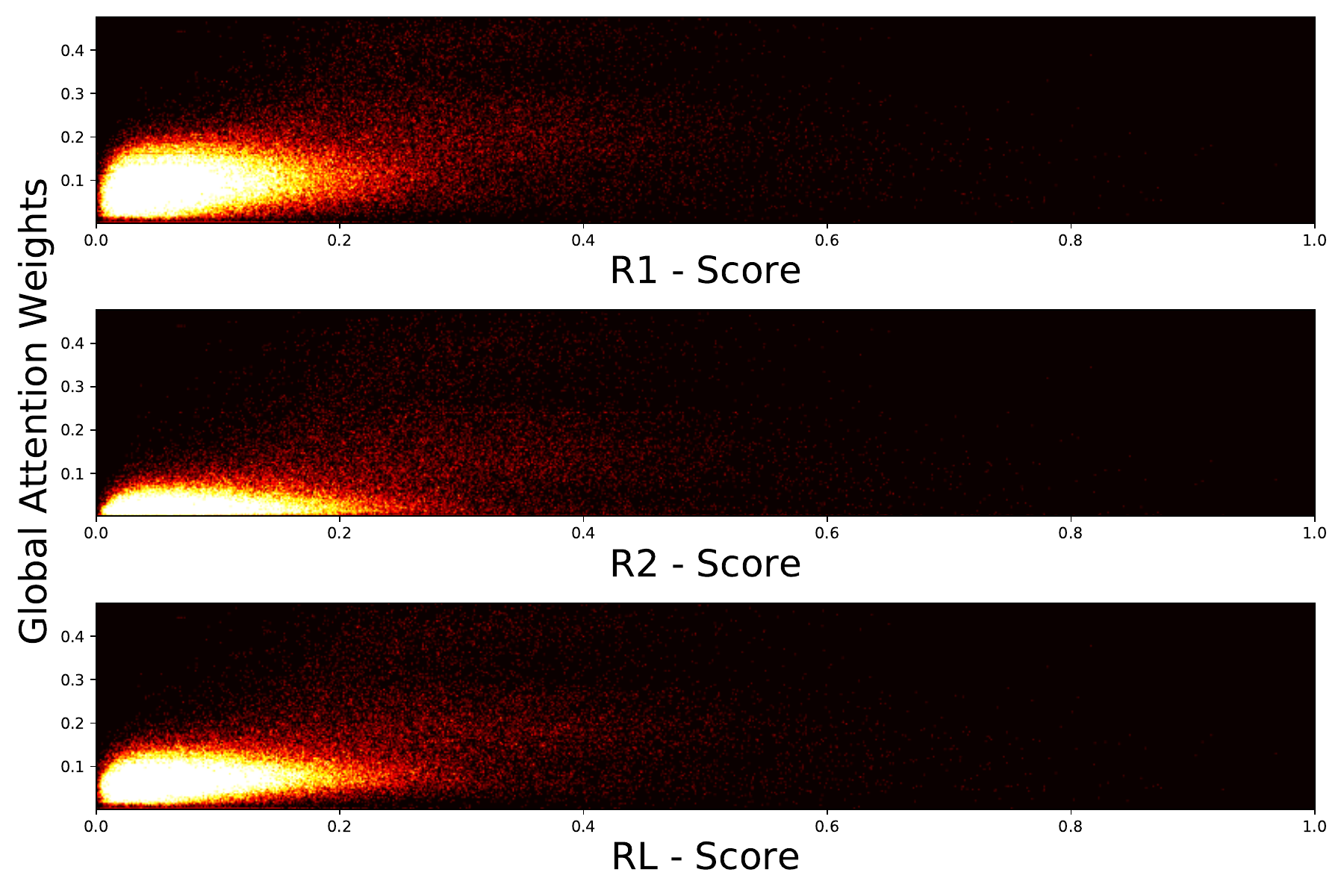}}

    \caption{Visualization of correlation between ROUGE scores indicating source origin information via text similarity and the aggregated global attention weights for the decoding layers not visualized in \autoref{fig:rq2-multinews-correlation}. A reduced test set is used for evaluation.  
    }
    \label{fig:suplementary-multinews-correlation}
\end{figure*}

\begin{figure*}[p]
\captionsetup[subfigure]{textfont=normal}
    \centering
    \subcaptionbox{\textbf{WikiSum} Decoding Layer 1}
        {\includegraphics[width=0.9\columnwidth]{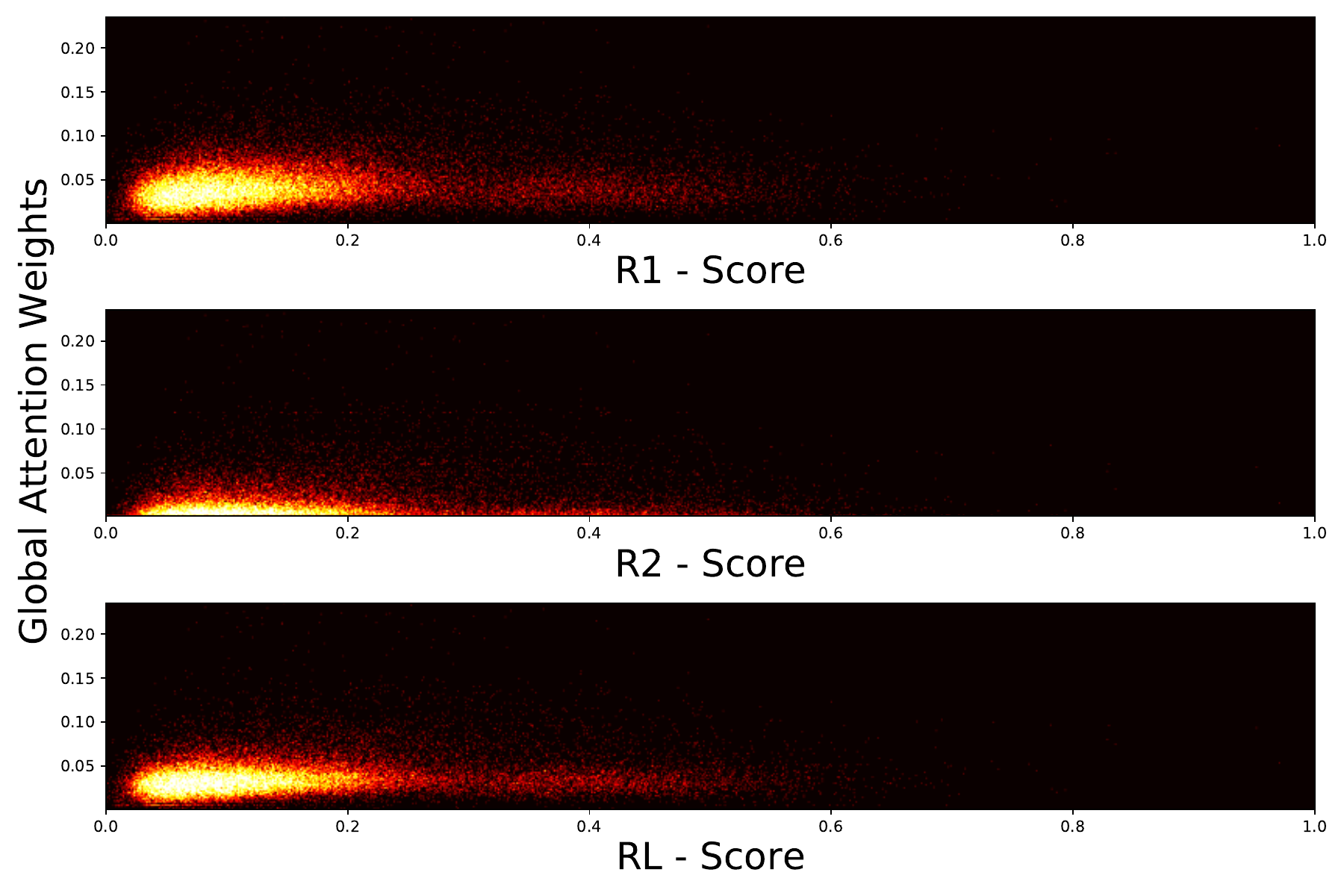}}
        \subcaptionbox{\textbf{WikiSum} Decoding Layer 2}
        {\includegraphics[width=0.9\columnwidth]{images/RQ2/wikisum_correlation/correlation_plot_layer_2.pdf}}
        \subcaptionbox{\textbf{WikiSum} Decoding Layer 3}
        {\includegraphics[width=0.9\columnwidth]{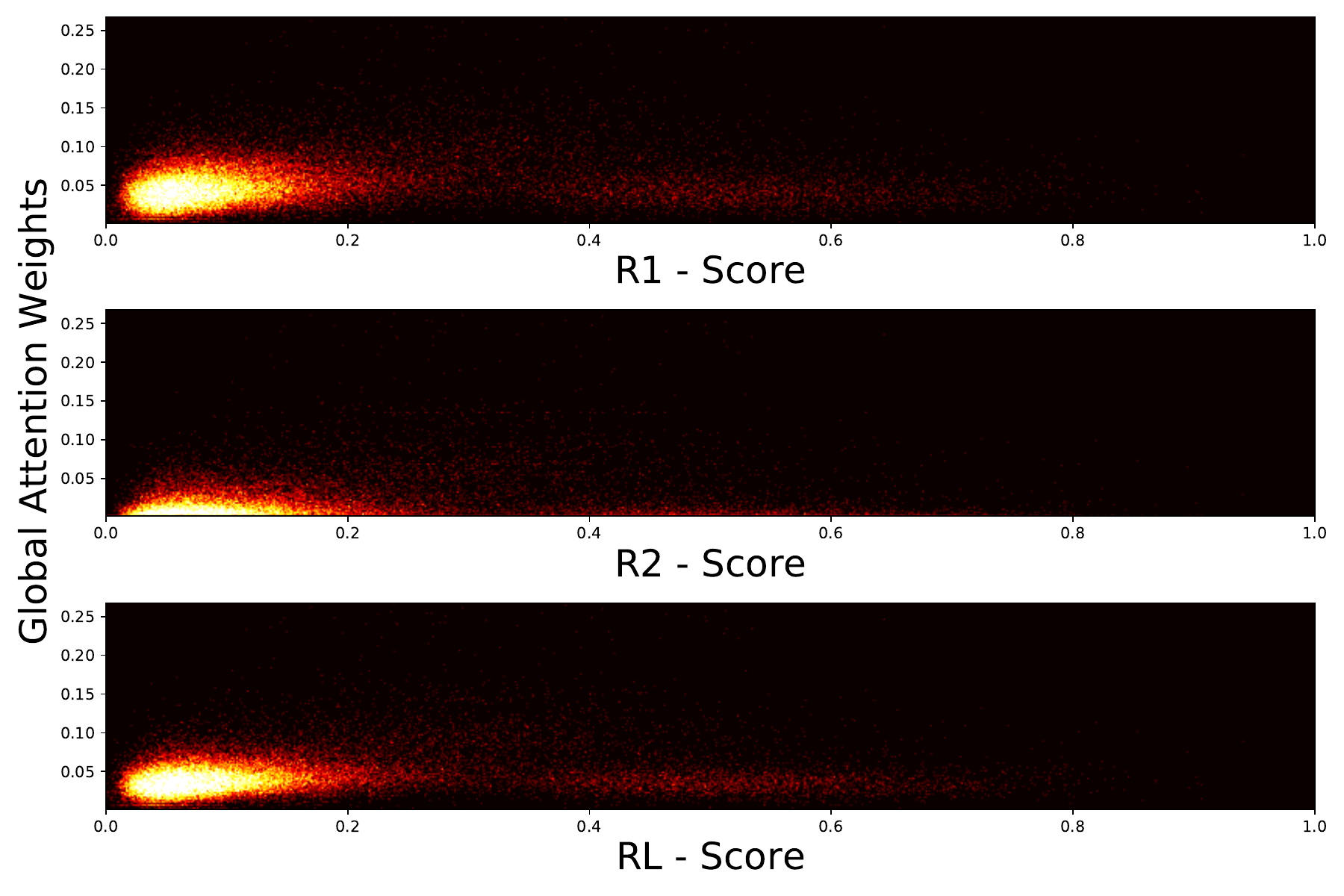}}
        \subcaptionbox{\textbf{WikiSum} Decoding Layer 4}
        {\includegraphics[width=0.9\columnwidth]{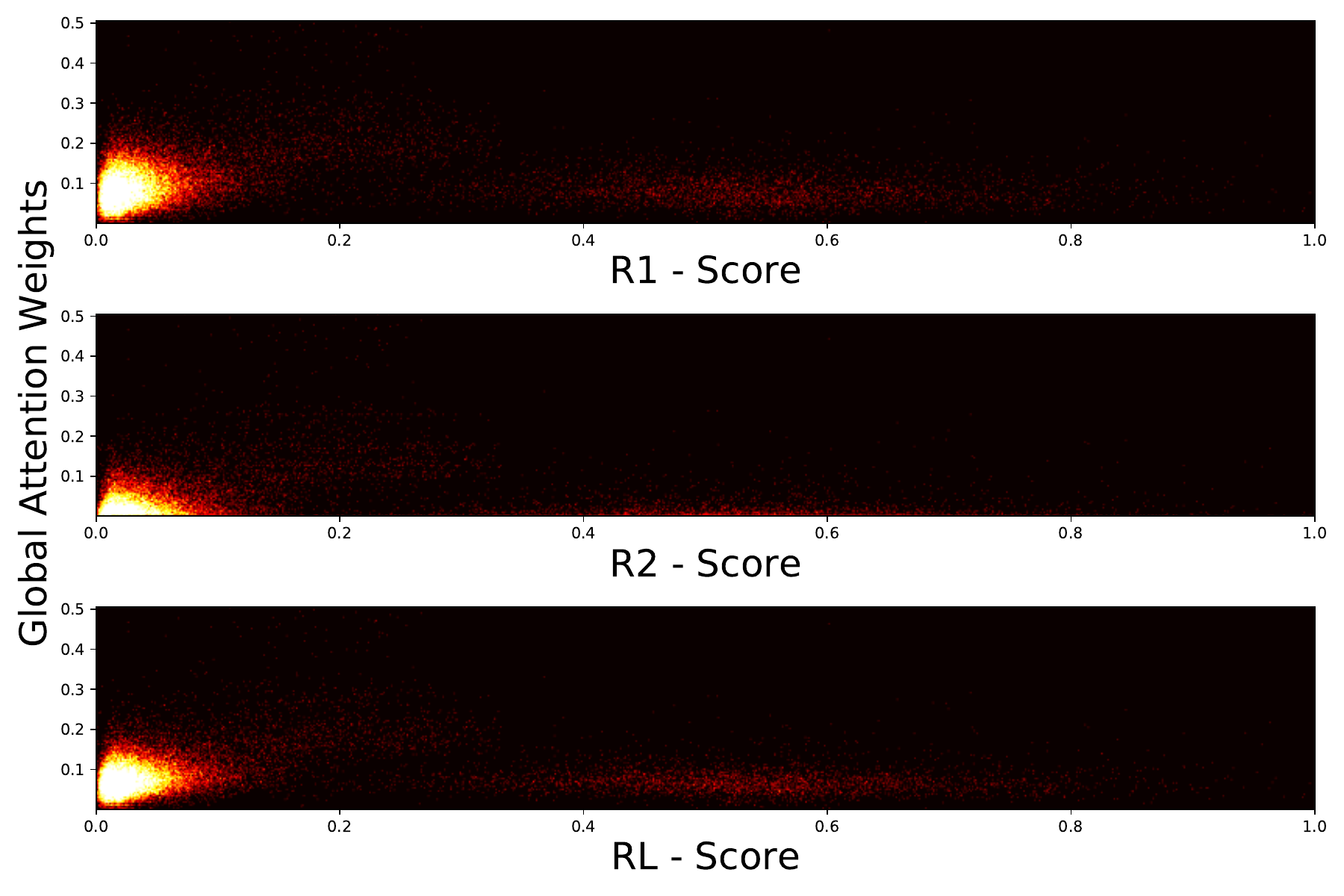}}
        \subcaptionbox{\textbf{WikiSum} Decoding Layer 5}
        {\includegraphics[width=0.9\columnwidth]{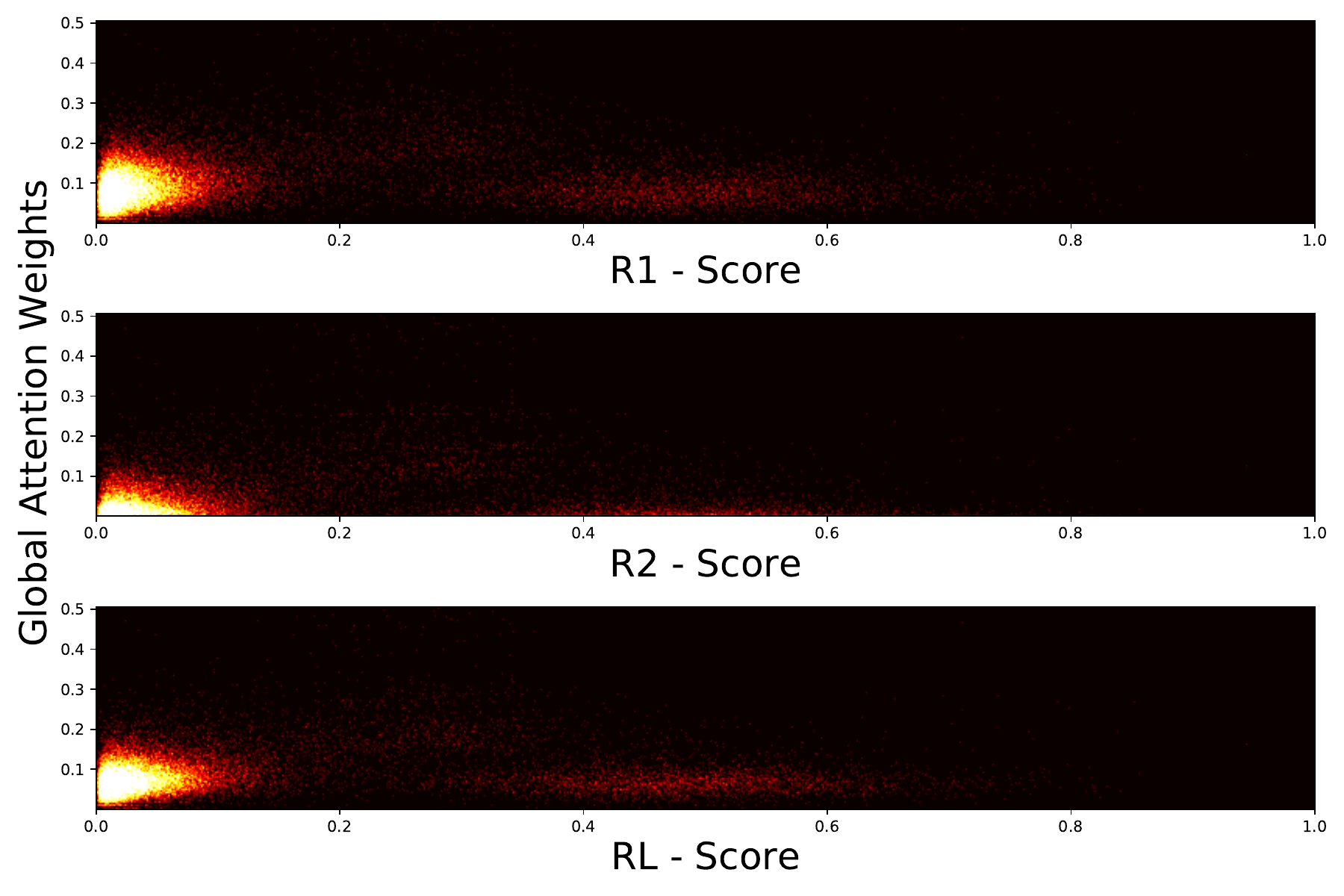}}
        \subcaptionbox{\textbf{WikiSum} Decoding Layer 6}
        {\includegraphics[width=0.9\columnwidth]{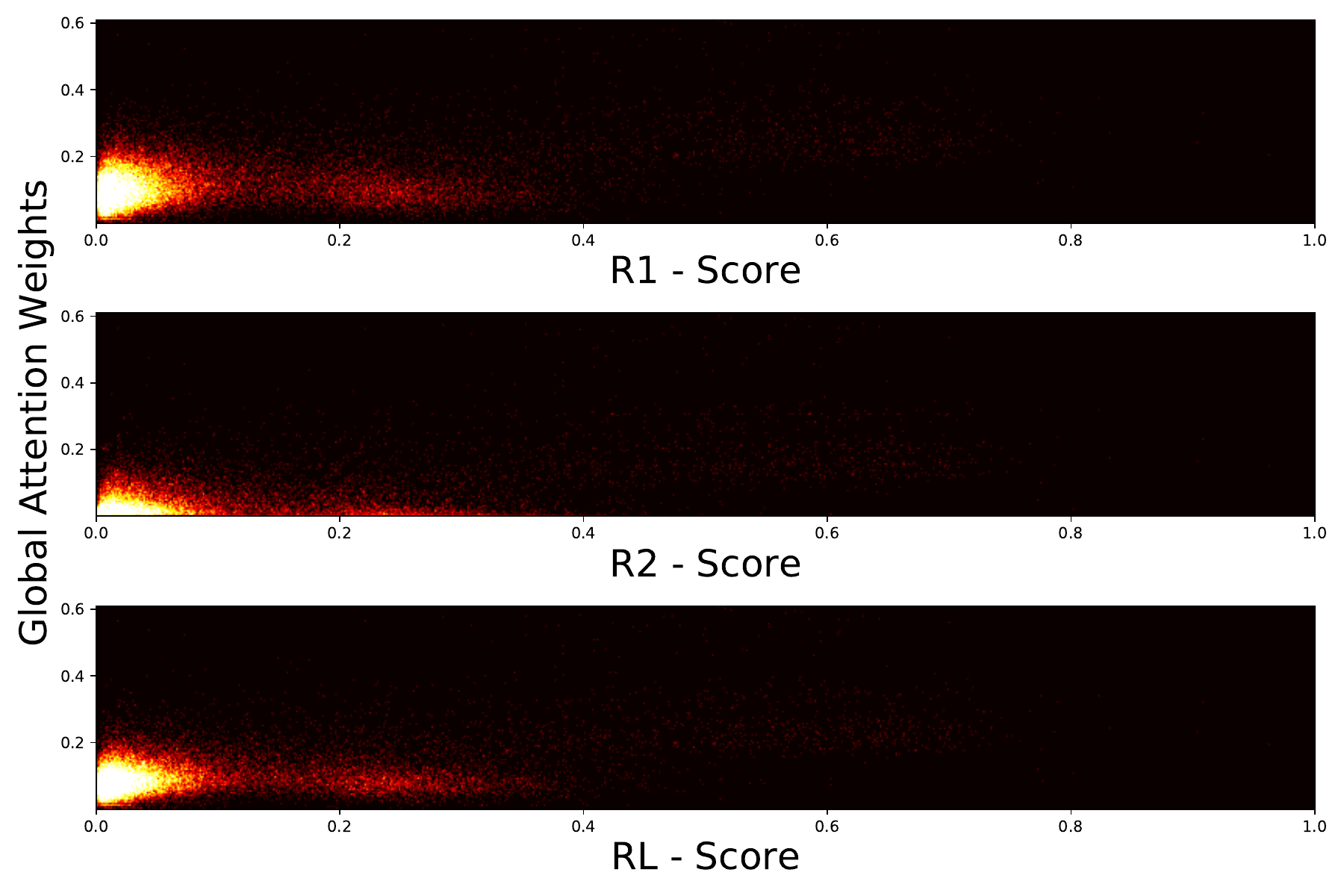}}
        \subcaptionbox{\textbf{WikiSum} Decoding Layer 7}
        {\includegraphics[width=0.9\columnwidth]{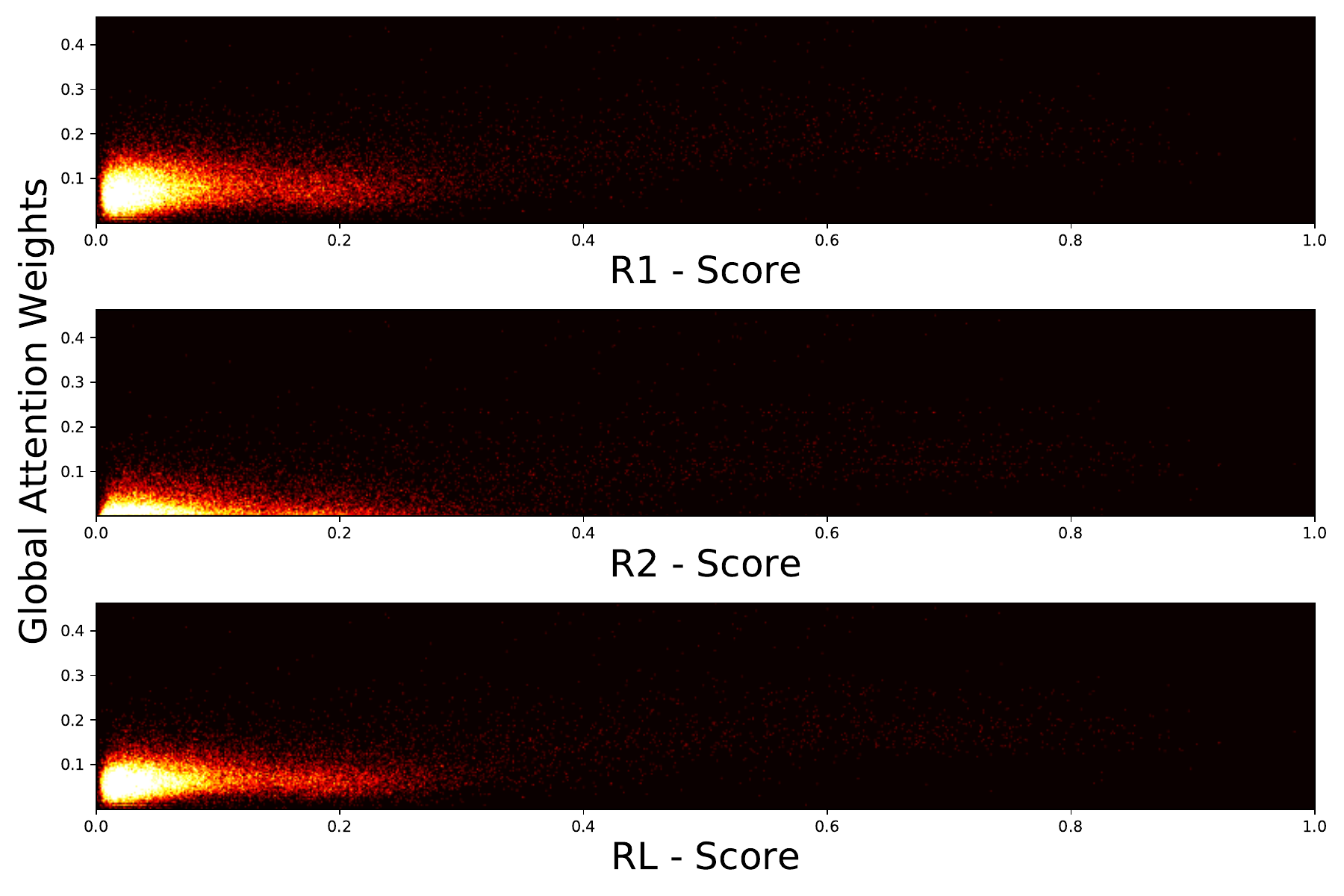}}
        \subcaptionbox{\textbf{WikiSum} Decoding Layer 8}
        {\includegraphics[width=0.9\columnwidth]{images/RQ2/wikisum_correlation/correlation_plot_layer_8.pdf}}
        \caption{Visualization of correlation between ROUGE scores indicating source origin information via text similarity and the aggregated global attention weights for the decoding layers not visualized in \autoref{fig:rq2-wiki-correlation}. A reduced test set is used for evaluation.  
    }
    \label{fig:suplementary-wiki-correlation}
\end{figure*}

The correlation plots for the decoding layers, not visualized in Section \ref{sec:results:RQ2} for MultiNews and WikiSum can be found in \autoref{fig:suplementary-multinews-correlation} and  \autoref{fig:suplementary-wiki-correlation}.

\section{Positional Bias}
\label{sec:appendix:positional_bias}

\begin{figure*}[p]
    \centering
    \subcaptionbox{Decoding Layer 1}
        {\includegraphics[width=0.78\columnwidth]{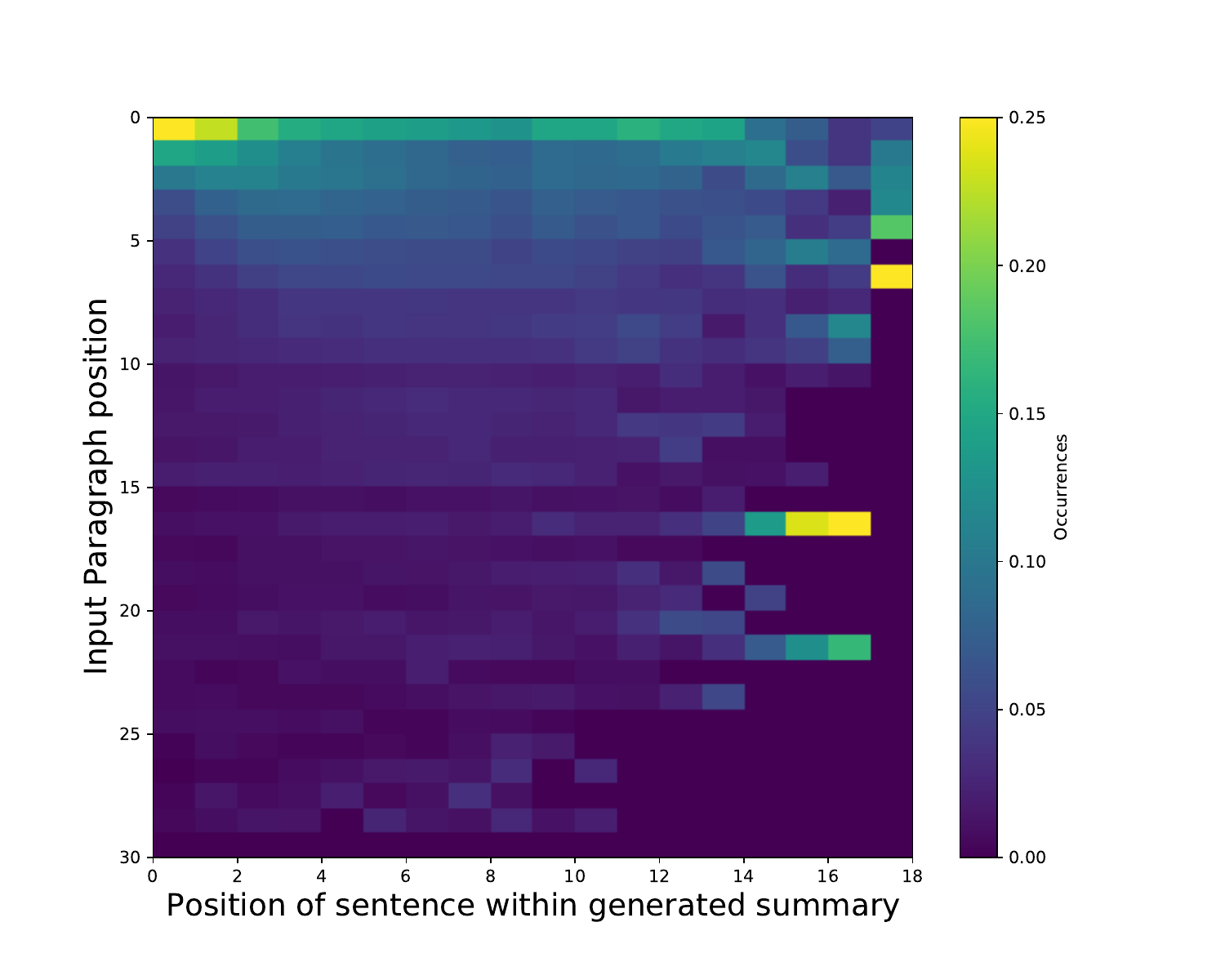}}
        \subcaptionbox{Decoding Layer 2}
        {\includegraphics[width=0.78\columnwidth]{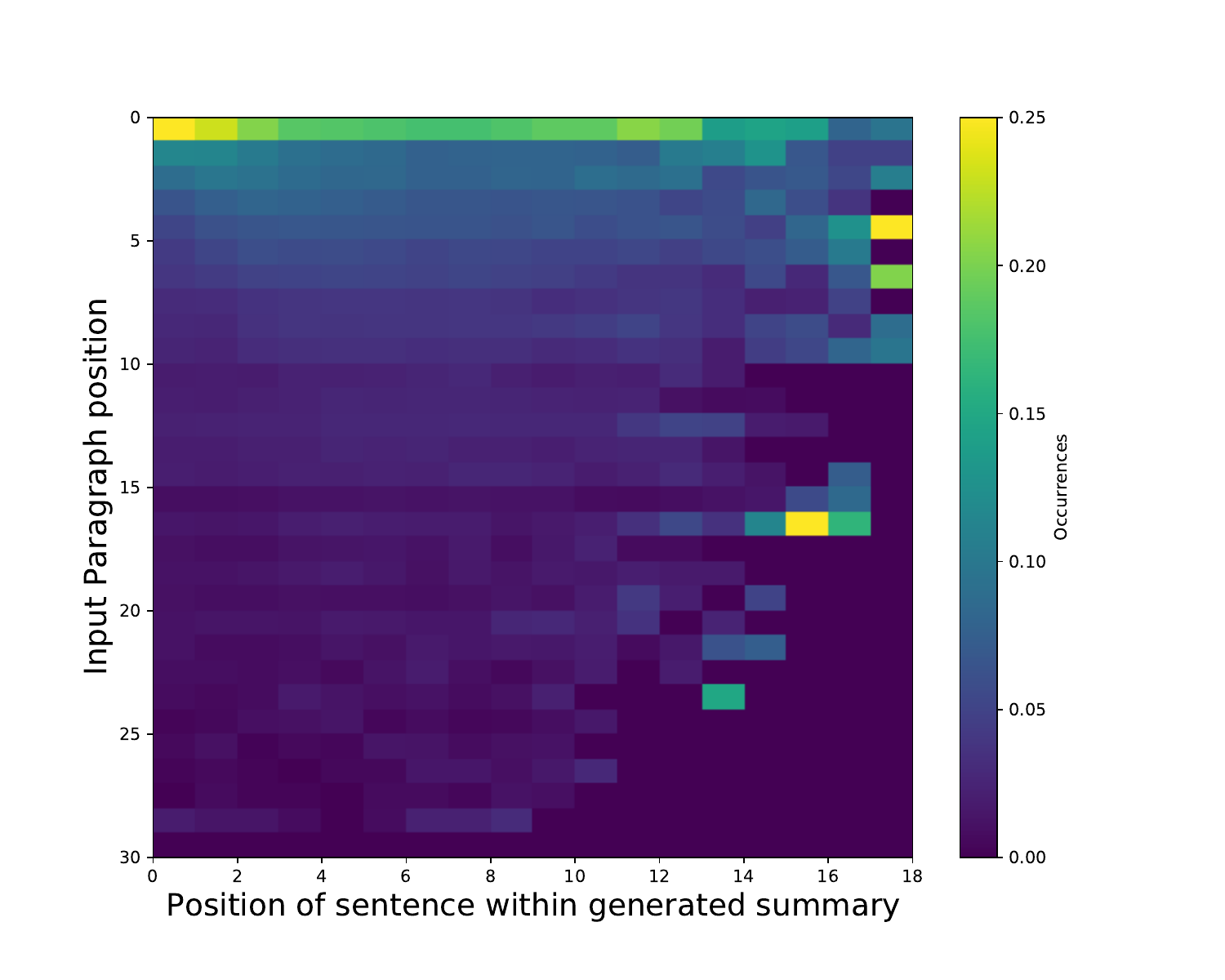}}
        \subcaptionbox{Decoding Layer 3}
        {\includegraphics[width=0.78\columnwidth]{images/RQ2/positional_bias/overall_norm_distri_dec_layer_3.pdf}}
        \subcaptionbox{Decoding Layer 4}
        {\includegraphics[width=0.78\columnwidth]{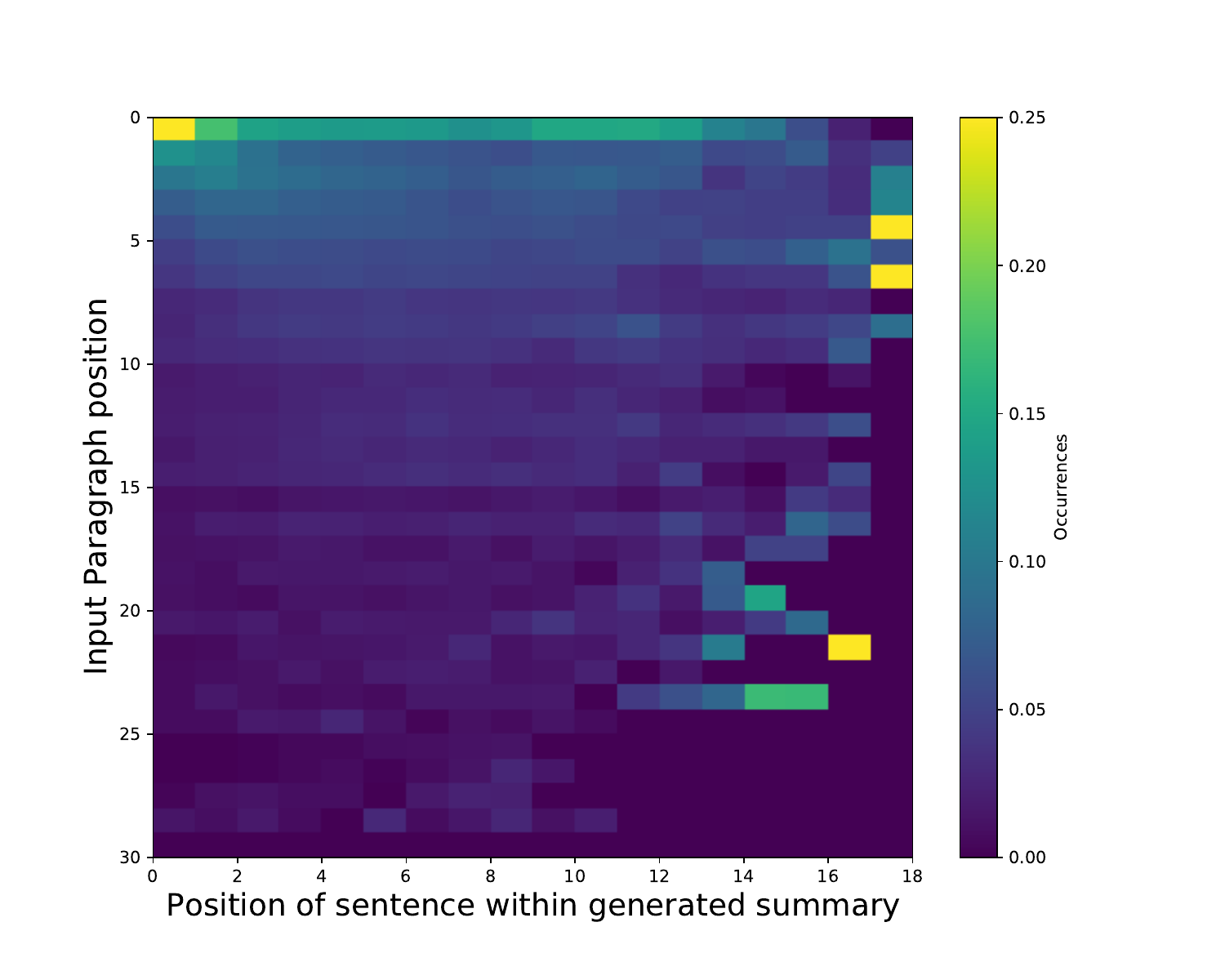}}
        \subcaptionbox{Decoding Layer 5}
        {\includegraphics[width=0.78\columnwidth]{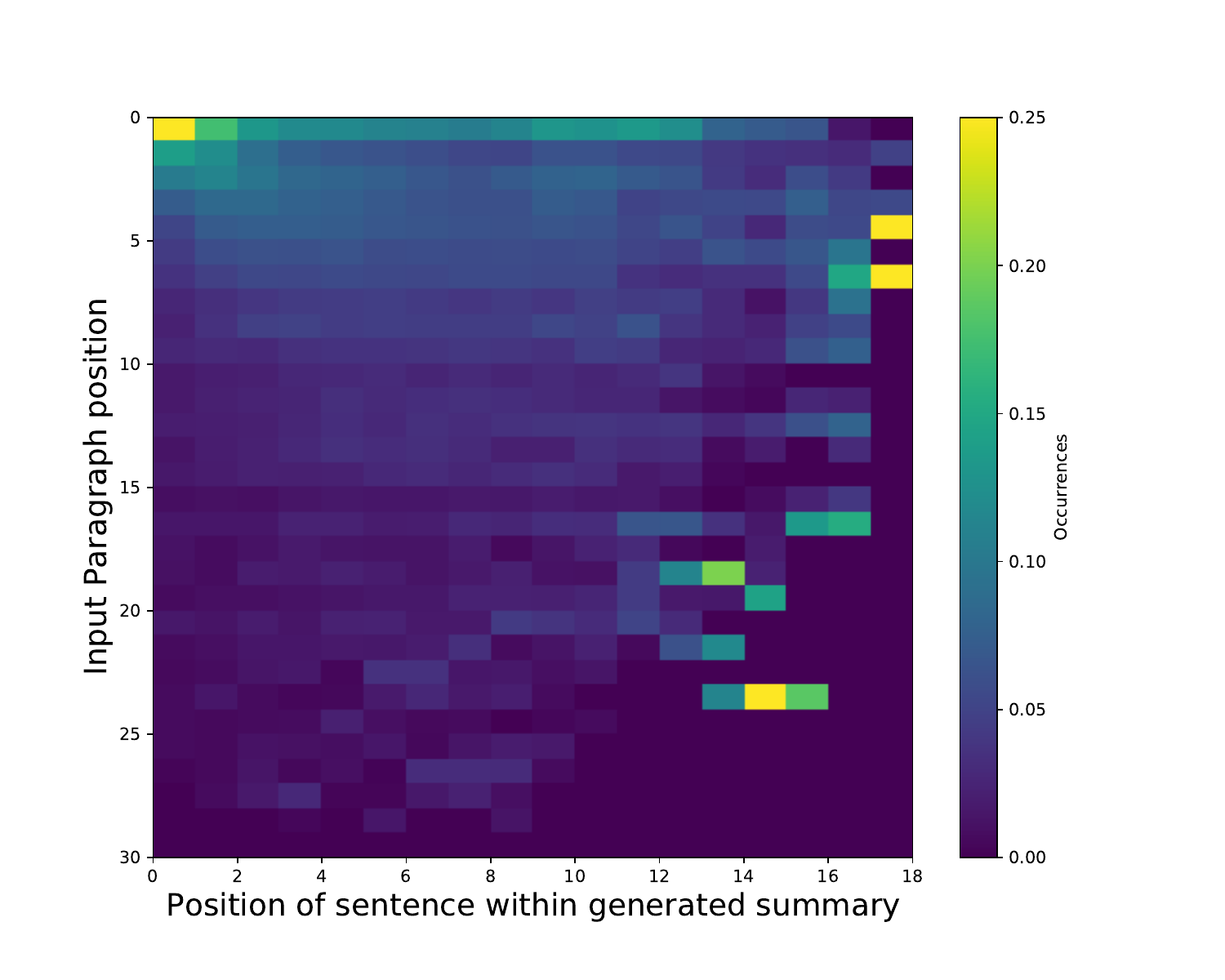}}
        \subcaptionbox{Decoding Layer 6}
        {\includegraphics[width=0.78\columnwidth]{images/RQ2/positional_bias/overall_norm_distri_dec_layer_6.pdf}}
        \subcaptionbox{Decoding Layer 7}
        {\includegraphics[width=0.78\columnwidth]{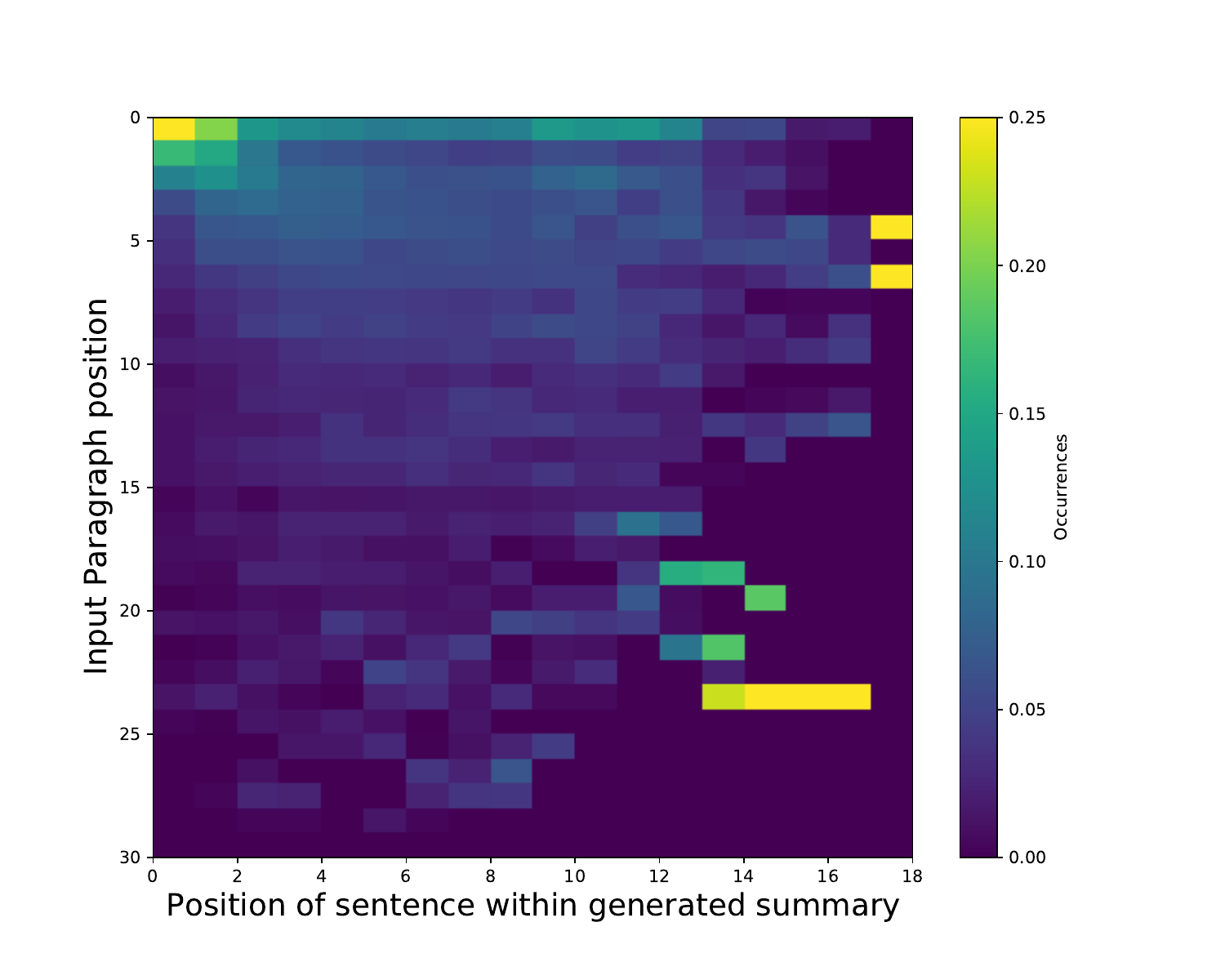}}
        \subcaptionbox{Decoding Layer 8}
        {\includegraphics[width=0.78\columnwidth]{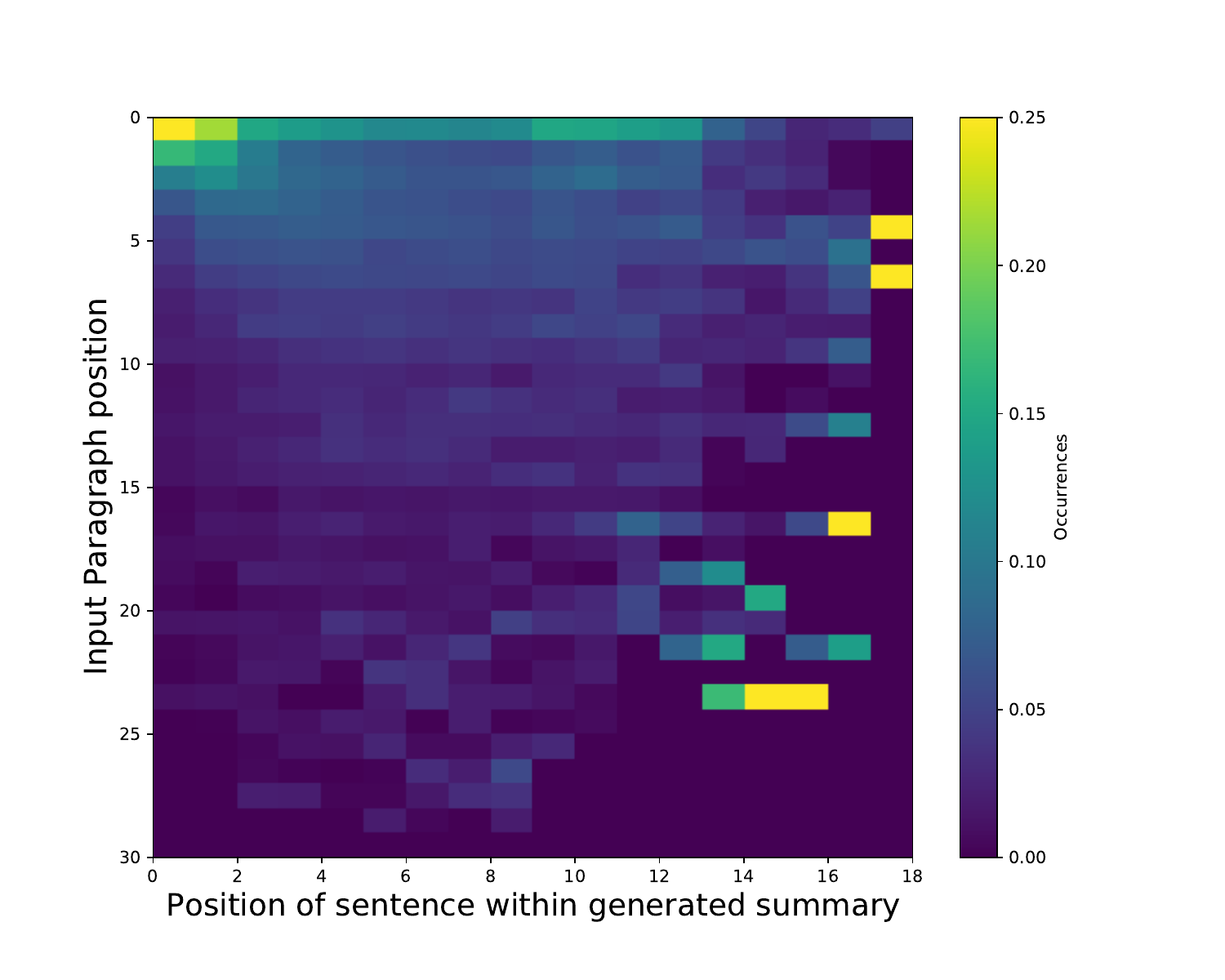}}

    \caption{Heatmap to visualize the paragraph positions, which were attended the most for each generated sentence. The analysis is performed on the MutliNews dataset.}
    \label{fig:suplementary-rq2-pos-bias-att}
\end{figure*}

The presence of positional bias in the generated summaries indicated by attention weights is shown for all decoding layers in \autoref{fig:suplementary-rq2-pos-bias-att}.

\end{document}